%% file: ablearn_arxiv.tex
\documentclass[10pt,onecolumn,letterpaper]{article}

\usepackage{times}
\usepackage{epsfig}
\usepackage{graphicx}
\usepackage{amsmath}
\usepackage{adjustbox}
\usepackage{amssymb}
\usepackage[vlined,boxed,commentsnumbered]{algorithm2e}
\usepackage{bm}
\usepackage{subfigure}
\usepackage[margin=0.5in]{geometry}
\usepackage{epsfig}

\input{definitions}
\def\MH#1{{\bf [Mehrtash:} {\it\color{blue} {#1}}{\bf ]}.}

\newcommand{\BlackBox}{\rule{1.5ex}{1.5ex}}  
\newenvironment{proof}{\par\noindent{\bf Proof\ }}{\hfill\BlackBox\\[2mm]}
\newtheorem{remark}{Remark}

\begin{document}

\title{Learning Discriminative $\alpha\beta$-divergence for Positive Definite Matrices\\ (Extended Version)\footnote{Extended version of paper accepted at the International Conference on Computer Vision (ICCV), Venice, 2017.}}

\author{$^1$A. Cherian\thanks{Equal contribution.}\quad$^2$P. Stanitsas$^\dagger$\quad$^3$M. Harandi\quad$^2$V. Morellas\quad$^2$N. Papanikolopoulos\\
$^1$Australian Centre for Robotic Vision, $^3$Data61/CSIRO, $^{1,3}$The Australian National University\\
$^2$Dept. of Computer Science, University of Minnesota, Minneapolis\\
\{anoop.cherian, mehrtash.harandi\}@anu.edu.au, \{stani078, morellas, npapas\}@umn.edu
}

\maketitle

\input{abstract}
\input{intro}
\input{related_work}
\input{background}

\input{proposed_method_arxiv}
\input{expts_arxiv}
\input{supp_mat}
\input{conclude}
\section*{Acknowledgments}This material is based upon work supported by the National Science Foundation through grants \#CNS-0934327, \#CNS-1039741, \#SMA-1028076, \#CNS-1338042, \#CNS-1439728, \#OISE-1551059, and \#CNS-1514626. Dr. Cherian is funded by the Australian Research Council Centre of Excellence for Robotic Vision (\#CE140100016).

{\small
\bibliographystyle{ieee}
\bibliography{CAD}
}

\end{document}

%% file: definitions.tex
\newtheorem{definition}{Definition}
\newtheorem{theorem}{Theorem}

\newcommand{\spd}[1]{\mathcal{S}_{++}^{#1}}

\newcommand{\eye}[1]{\mathbf{I}_{d}}

\newcommand{\reals}[1]{\mathbb{R}^{#1}}

\newcommand{\half}{\frac{1}{2}}

\newcommand{\comment}[1]{}

\newcommand{\ab}{\alpha\beta}

\newcommand{\abldab}[4]{D^{({#3},{#4})}({#1}\parallel{#2})}
\newcommand{\logdet}{\log\det}
\DeclareMathOperator*{\Log}{Log}
\DeclareMathOperator*{\Exp}{Exp}
\newcommand\inv[1]{#1\raisebox{1.15ex}{$\scriptscriptstyle\!\!-\!1$}}
\newcommand{\set}[1]{\left\{#1\right\}}
\newcommand{\enorm}[1]{\left\|{#1}\right\|}
\newcommand{\trace}[1]{Tr\left({#1}\right)}
\newcommand{\grad}[1]{\nabla_{#1}}
\newcommand{\fnorm}[1]{\left\|{#1}\right\|_F}

\DeclareMathOperator*{\diag}{diag}

\newcommand{\bigoh}{\mathcal{O}}
\DeclareMathOperator*{\gridsearch}{Grid Search}
\DeclareMathOperator*{\kmeans}{kmeans}

\newcommand{\mX}{X}
\newcommand{\mY}{Y}
\newcommand{\mA}{A}
\newcommand{\vx}{\mathbf{x}}

\newcommand{\sx}{x}
\newcommand{\salpha}{\alpha}
\newcommand{\sbeta}{\beta}
\newcommand{\dataset}{\mathcal{X}}
\newcommand{\labels}{\mathcal{L}}
\newcommand{\dict}{\mathbf{B}}
\newcommand{\mB}{B}
\newcommand{\vv}{\mathbf{v}}
\newcommand{\valpha}{\bm{\alpha}}
\newcommand{\vbeta}{\bm{\beta}}
\newcommand{\mW}{W}
\newcommand{\sy}{y}
\newcommand{\vh}{\mathbf{h}}
\newcommand{\vr}{\mathbf{r}}

\newcommand{\mZ}{Z}
\newcommand{\vtheta}{\mathbf{\theta}}
\newcommand{\mH}{H}
\newcommand{\mV}{V}
\newcommand{\mU}{U}

\newcommand{\mDelta}{\Delta}
\newcommand{\vdelta}{\bm{\delta}}
\newcommand{\mS}{S}

\newcommand{\tb}[1]{\textbf{#1}}

\newcommand{\iddl}{\text{IDDL}}

%% file: abstract.tex
\begin{abstract}
Symmetric positive definite (SPD) matrices are useful for capturing second-order statistics of visual data. To compare two SPD matrices, several measures are available, such as the affine-invariant Riemannian metric, Jeffreys divergence, Jensen-Bregman logdet divergence, etc.; however, their behaviors may be application dependent, raising the need of manual selection to achieve the best possible performance. Further and as a result of their overwhelming complexity for large-scale problems, 
computing pairwise similarities by clever embedding of SPD matrices is often preferred
to direct use of the aforementioned measures. In this paper, we propose a discriminative metric learning framework,~\emph{Information Divergence and Dictionary Learning} (IDDL), that not only learns application specific measures on SPD matrices \emph{automatically}, but also embeds them as vectors using a learned dictionary. To learn the similarity measures (which could potentially be distinct for every dictionary atom), we use the recently introduced $\alpha\beta$-logdet divergence, which is known to unify the measures listed above. We propose a novel IDDL objective, that learns the parameters of the divergence and the dictionary atoms jointly in a discriminative setup and is solved efficiently using Riemannian optimization. We showcase extensive experiments on eight computer vision datasets, demonstrating state-of-the-art performances.

\end{abstract}

%% file: intro.tex
\section{Introduction}
\label{sec:intro}
Symmetric Positive Definite (SPD) matrices arise naturally in several computer vision applications, such as covariances when modeling data using Gaussians, as kernel matrices for high-dimensional embedding, as points in diffusion MRI~\cite{pennec2006}, and as structure tensors in image processing~\cite{brox2006nonlinear}. 
Furthermore, SPD matrices in the form of Region CoVariance Descriptors (RCoVDs) ~\cite{tuzel2006region}, 
offer an easy way to compute a representation that fuses multiple modalities (e.g., color, gradients, filter responses, etc.) in a cohesive,  and compact format. In various mainstream vision applications, including tracking, re-identification, object, texture, and activity recognition,  the trail of SPD matrices to advance the state-of-the-art solutions can be seen~\cite{carreira2012semantic,wang2015beyond,harandi2015riemannian}. SPD matrices are even used as second-order pooling operators for enhancing the performance of popular deep learning architectures~\cite{ionescu2015matrix,huang2016riemannian}.

\begin{figure}[t]
\begin{center}
\includegraphics[width=12cm,bb=0 -1 960 540]{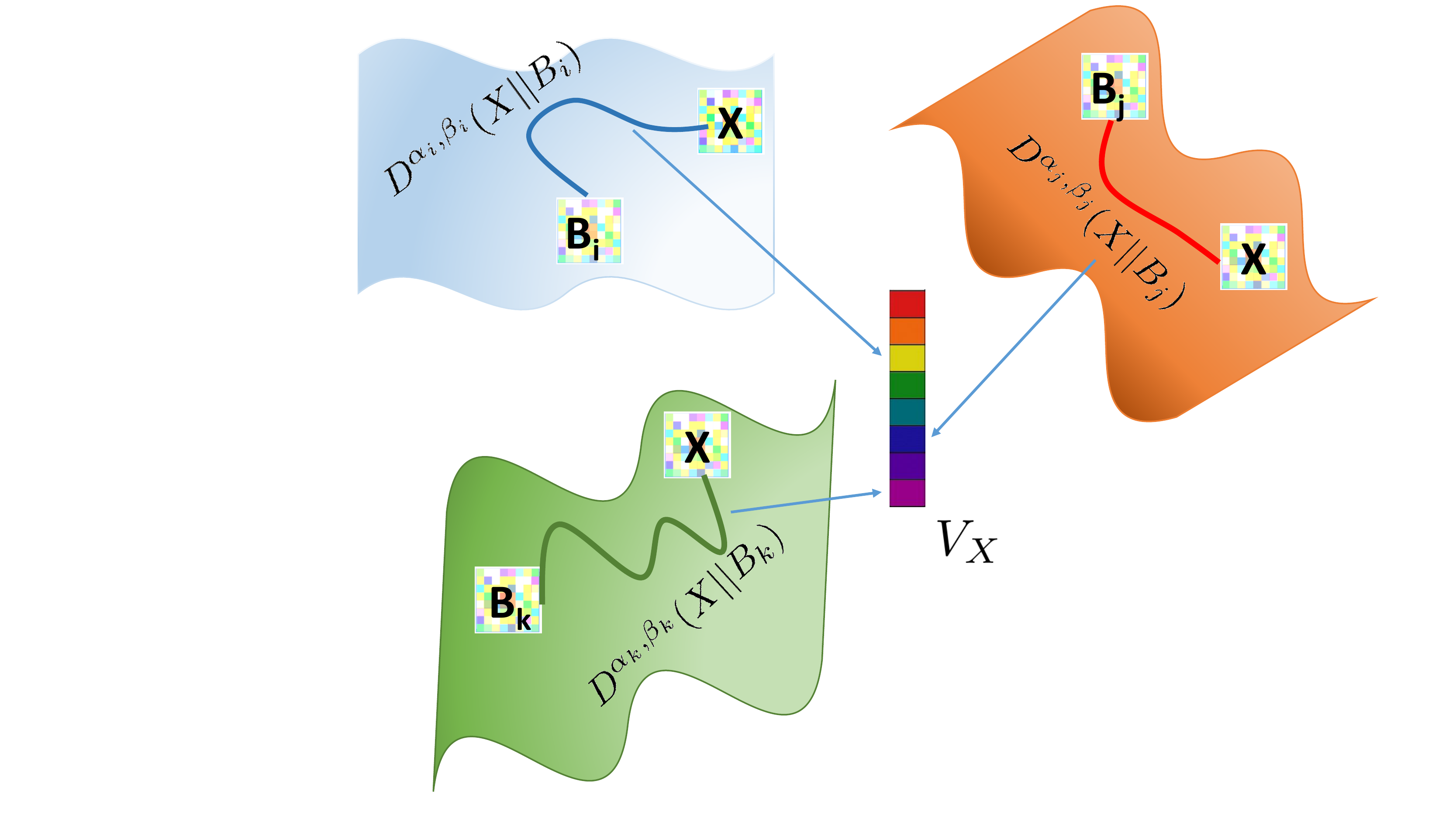}
\caption{A schematic illustration of our IDDL scheme. From an infinite set of potential geometries, our goal is to learn multiple geometries (parameterized by $(\alpha,\beta)$) and representative dictionary atoms for each geometry (represented by $B$'s), such that a given SPD data matrix $X$ can be embedded into a similarity vector $V_X$, each dimension of which captures the divergence of $X$ to the $B$s using the respective measure. We use $V_X$ for classification.}
\end{center}
\end{figure}

SPD matrices, due to their positive definiteness property, form a cone in the Euclidean space. However, 
analyzing these matrices through their Riemannian geometry (or the associated Lie algebra) 
helps avoiding unlikely/unrealistic solutions, thereby improving the outcomes. For example, in diffusion MRI~\cite{pennec2006,arsigny2006log}, it has been shown that the Riemannian structure (which comes with an affine invariant metric) is immensely useful for accurate modeling. A similar observation is made for RCoVDs~\cite{cherian2013jensen,harandi2014manifold,wang2012covariance}. This has resulted in the exploration of various geometries and similarity measures for SPD matrices, viewing them from disparate perspectives. A few notable such measures are: (i) the affine invariant Riemannian metric (AIRM) using the natural Riemannian geometry~\cite{pennec2006}, (ii) the Jeffreys KL divergence (KLDM) using relative entropy~\cite{moakher2006symmetric}, (iii) the Jensen-Bregman logdet divergence using information geometry~\cite{cherian2013jensen}, and (iv) Brug matrix divergence~\cite{kulis2006learning}, among several others~\cite{cichocki2015log}.

Each of the aforementioned measures has distinct mathematical properties and as such performs differently for a given problem. 
However and to some extent surprisingly, all of them  can be obtained as functions acting on the generalized eigenvalues of their inputs. Recently, Cichocki et al.~\cite{cichocki2015log} show that all these measures can be interpreted in a unifying setup using $\alpha\beta$-logdet divergence (ABLD) and each measure can be derived as a distinct parametrization of this divergence. For example, one could get JBLD from ABLD\footnote{Up to a scaling factor.} using $\alpha=\beta=\frac{1}{2}$, and AIRM as the limit of $\alpha,\beta\rightarrow 0$. 
With such an interesting discovery, it is natural to ask if the parameters $\alpha$ and $\beta$ can be learned for a given task 
in a data-driven way. \emph{This not only answers which measure is the right choice for a given problem, but also allows for deriving new measures  that are not among the popular ones listed above. }

In this paper, we make the first attempt at learning an $\alpha\beta$-logdet divergence on SPD matrices for computer vision applications, dubbed~\emph{Information Divergence and Dictionary Learning} (IDDL). We cast the learning problem in a discriminative ridge regression setup where the goal is to learn $\alpha$ and $\beta$ that maximize the classification accuracy for a given task. 

Being vigilant to the computational complexity of the resulting solution, we propose to embed SPD matrices using a dictionary in our metric learning framework. Our proposal enables us to learn the embedding (or more accurately the dictionary that identifies the embedding),
along with the proper choice of the metric (i.e., parameters $\alpha$ and $\beta$ of the ABLD) and a classifier jointly. 
The output of our IDDL is a vector, each entry of this vector computes a potentially distinct ABLD to a distinct dictionary atom.

To achieve our goal, we propose an efficient formulation that benefits from recent advances in optimization over Riemannian manifolds to minimize a non-convex and constrained objective.
%
%
%
%
We provide extensive experiments using IDDL on a variety of computer vision applications, namely (i) action recognition, (ii) texture recognition, (iii) 3D shape recognition, and (iv) cancerous tissue recognition. We also provide insights into our learning scheme 
through extensive experiments on the parameters of the ABLD, and ablation studies under various performance settings. Our results demonstrate that our scheme achieves state-of-the-art accuracies against competing techniques, including the recent sparse coding, Riemannian metric learning, and kernel coding schemes.

%% file: related_work.tex
\section{Related Work}
\label{sec:related_work}
The $\ab$-logdet divergence is a matrix generalization of the well-known $\ab$-divergence~\cite{cichocki2010families} that computes the (a)symmetric (dis)similarity between two finite positive measures (data densities). As the name implies, $\ab$-divergence is a unification of the so-called $\alpha$-family of divergences~\cite{amari2007methods} (that includes popular measures such as the KL-divergence, Jensen-Shannon divergence, and the chi-square divergence) and the $\beta$-family~\cite{basu1998robust} (including the squared Euclidean distance and the Itakura Saito distance). Against several standard measures for computing similarities, both $\alpha$ and $\beta$ divergences are known to lead to solutions that are robust to outliers and additive noise~\cite{lafferty1999additive}, thereby improving application accuracy. They have been used in several statistical learning applications including non-negative matrix factorization
~\cite{cichocki2009nonnegative,kompass2007generalized,dhillon2005generalized}, nearest neighbor embedding~\cite{hinton2002stochastic}, and blind-source separation~\cite{mihoko2002robust}.  

A class of methods with similarities to our formulation are metric learning schemes on SPD matrices. One popular technique is the manifold-manifold embedding of large SPD matrices into a tiny SPD space in a discriminative setting~\cite{harandi2014manifold}. Log-Euclidean metric learning has also been proposed for this embedding in~\cite{huang2015log,sivalingam2009metric}. While, we also learn a metric in a discriminative setup, ours is different in that we learn an information divergence. In Thiyam et al.~\cite{thiyam2017optimization}, ABLD is proposed replacing symmetric KL divergence in better characterizing the learning of a decision hyperplane for BCI applications. In contrast\footnote{Automatic selection of the parameters of $\ab$-divergence is investigated in~\cite{csimcsekli2015learning,dikmen2015learning}. However, they deal with scalar density functions in a maximum-likelihood setup and do not consider the optimization of $\alpha$ and $\beta$ jointly.}, we propose to embed the data matrices as vectors, each dimension of these vectors learning a different ABLD, thus leading to a richer representation of the input matrix. 

Vectorial embedding of SPD matrices has been investigated using disparate formulations for computer vision applications. As alluded to earlier, the log-Euclidean projection~\cite{arsigny2006log} is a common way to achieve this, where an SPD matrix is isomorphically mapped to the Euclidean space of symmetric matrices using the matrix logarithm. Popular sparse coding schemes have been extended to SPD matrices in~\cite{cherian2016riemannian, sivalingam2010tensor,xie2013nonlinear}
 using SPD dictionaries, where the resulting sparse vector is assumed Euclidean. Another popular way to handle the non-linear geometry of SPD matrices is to resort to kernel schemes by embedding the matrices in an infinite dimensional Hilbert space which is assumed to be linear ~\cite{harandi2014bregman,li2013log,harandi2015riemannian}. In all these methods, the underlying similarity measure is fixed and is usually chosen to be one among the popular $\ab$-logdet divergences or the log-Euclidean metric. 
 
 In contrast to all these methods, to the best of our knowledge, it is for the first time that a joint dictionary learning and information divergence learning framework is proposed for SPD matrices in computer vision. In the sequel, we first introduce $\ab$-logdet divergence and explore its properties in the next section. This will precede exposition to our discriminative metric learning framework for learning the divergence and efficient ways of solving our formulation. 
 
 \noindent\paragraph{Notations:} Following standard notations, we use upper case for matrices (such as $\mX$), lower-bold case for vectors $\vx$, and lower case for scalars $\sx$. Further, $\spd{d}$ is used to denote the cone of $d\times d$ SPD matrices. We use $\dict$ to denote a 3D tensor each slice of which is an SPD matrix of size $d\times d$. Further, we use $\eye{d}$ to denote the $d\times d$ identity matrix, $\Log$ for the matrix logarithm, and $\diag$ for the diagonalization operator.
 

%% file: background.tex
\section{Background}
\label{sec:background}
In this section, we will setup the mathematical preliminaries necessary to elucidate our contributions. We will visit the $\ab$-log-det divergence, its connections to other popular divergences, and its mathematical properties.

\begin{table*}[htbp]
\centering
\begin{tabular}{|c|c|c|}
\hline
$(\alpha,\beta)$  & ABLD & Divergence\\
\hline
$(\alpha,\beta)\rightarrow 0$ & $\fnorm{\Log{\mX^{-\half}\mY\mX^{-\half}}}^2$ &  Squared Affine Invariant Riemannian Metric~\cite{pennec2006}\\
\hline
$\alpha=\beta=\pm\half$ &  $4\left(\logdet\frac{\mX+\mY}{2}-\half\logdet\mX\mY\right)$ & Jensen-Bregman Logdet Divergence~\cite{cherian2013jensen} \\
\hline
$\alpha=\pm 1,\beta\rightarrow 0$ & $\half\trace{\mX\inv{\mY} + \mY\inv{\mX}} - d$ & Jeffreys KL Divergence\footnote{using the symmetrization of ABLD.}~\cite{moakher2006symmetric}\\
\hline
$\alpha=1,\beta=1$ & $\trace{\mX\inv{\mY}} - \logdet\mX\inv{\mY} - d$ & Burg Matrix Divergence~\cite{kulis2006learning} \\
\hline
\end{tabular}
\caption{ABLD and its connections to popular divergences used in computer vision applications.}
\label{tab:1}
\end{table*}

\subsection{$\alpha\beta$-Log Determinant Divergence}
\begin{definition}[ABLD~\cite{cichocki2015log}]
For $X,Y\in\spd{d}$, the $\ab$-log-det divergence is defined as:
\begin{equation}
	D^{(\alpha,\beta)}\hspace*{-0.3ex}(\mX \hspace*{-0.5ex} \parallel \hspace*{-0.5ex} \mY)
	\hspace*{-0.5ex} = \hspace*{-0.5ex} \frac{1}{\salpha\sbeta}\logdet \hspace*{-0.5ex}
	\left(\frac{\salpha(\mX\mY^{-1})^{\sbeta} \hspace*{-0.5ex} + \hspace*{-0.5ex}\sbeta(\mX\mY^{-1})^{-\salpha}}
	{\salpha+\sbeta}\right)\hspace*{-0.5ex},
	\label{eq:abld}	
\end{equation}
\begin{equation}
     \salpha\neq 0,\ \sbeta\neq 0\ \text{ and } \salpha+\sbeta\neq 0.
     \label{eq:abld_constraints}
\end{equation}
\end{definition}

It can be shown that ABLD depends only on the generalized eigenvalues of $\mX$ and $\mY$~\cite{cichocki2015log}. 
Suppose $\lambda_i$ denotes the $i$-th eigenvalue of $\mX\inv{\mY}$. Then under constraints defined in~\eqref{eq:abld_constraints}, 
we can rewrite~\eqref{eq:abld} as:
\begin{equation}
	D^{(\alpha,\beta)}\hspace*{-0.3ex}(\mX \hspace*{-0.7ex} \parallel \hspace*{-0.5ex} \mY \hspace*{-0.25ex})
    \hspace*{-0.5ex} = \hspace*{-0.5ex} \frac{1}{\salpha\sbeta} \hspace*{-0.5ex} \sum_{i=1}^d 
    \hspace*{-0.25ex} \log \Big(\salpha\lambda_i^{\sbeta} 
    \hspace*{-0.5ex} + \hspace*{-0.3ex} \sbeta \lambda_i^{-\salpha} 
    \Big)
    \hspace*{-0.25ex} - \hspace*{-0.25ex} d\log\left(\salpha \hspace*{-0.5ex} + \hspace*{-0.5ex} \sbeta \right)\hspace*{-0.5ex}.
    \label{eq:abld_lambda}
\end{equation}
 This formulation will come handy when deriving the gradient updates for $\alpha$ and $\beta$ in the sequel. As alluded to earlier, a hallmark of the ABLD is that it unifies several popular distance measures on SPD matrices that one commonly encounters in computer vision applications. In Table~\ref{tab:1}, we list some of the popular measures in computer vision and the respective values of $\alpha$ and $\beta$.

\subsection{ABLD Properties}
\noindent\textbf{Avoiding Degeneracy:} An important observation regarding the design of optimization algorithms on ABLD is that the quantity inside the $\logdet$ term has to be positive definite; conditions on $\alpha$ and $\beta$ for which are specified by the following theorem.
\begin{theorem}[\cite{cichocki2015log}]
For $\mX,\mY\in\spd{d}$, if $\lambda_i$ is the $i$-th eigenvalue of $\inv{\mX}\mY$, then $\abldab{\mX}{\mY}{\salpha}{\sbeta}\geq 0$ only if 
\begin{align}
\lambda_i & > \left|\frac{\alpha}{\beta}\right|^{\frac{1}{\alpha+\beta}}\!\!\!, \text{ for } \alpha>0 \text{ and } \beta < 0, \text{ or } \\
\lambda_i &< \left|\frac{\beta}{\alpha}\right|^{\frac{1}{\alpha+\beta}}\!\!\!,\text{ for } \alpha<0 \text{ and } \beta > 0, \forall i=1,2,\cdots, d.
\end{align}
\label{thm:1}
\end{theorem}

Since $\lambda_i$s depend on the input matrices, on which we have no control over, we constrain $\salpha$ and $\sbeta$ to have the same sign, thereby avoiding the quantity inside $\logdet{}$ to be indefinite. We make this assumption in our formulations in Section~\ref{sec:proposed_method}.

\noindent\textbf{Smoothness of $\alpha$, $\beta$:}
Assuming $\alpha,\beta$ have the same sign, except at origin ($\alpha=\beta=0$), ABLD is smooth everywhere with respect to $\alpha$ and $\beta$, thus allowing us to develop Newton-type algorithms on them. Due to the discontinuity at the origin, we ought to design algorithms specifically addressing this particular case.

\noindent\textbf{Affine Invariance:}
It can be easily shown that
\begin{equation}
\abldab{\mX}{\mY}{\salpha}{\sbeta} = \abldab{\mA\mX\mA^T}{\mA\mY\mA^T}{\salpha}{\sbeta},
\end{equation}
for any invertible matrix $\mA$. This is an important property that makes this divergence useful in a variety of applications, such as diffusion MRI~\cite{pennec2006}.

\noindent\textbf{Dual Symmetry:}
This property allows us to extend results derived for the case of $\alpha$  to the one on $\beta$ later.
\begin{equation}
\abldab{\mX}{\mY}{\salpha}{\sbeta} = \abldab{\mY}{\mX}{\sbeta}{\salpha}.
\label{eq:dual_sym}
\end{equation}
 
Before concluding this part, we briefly introduce the concept of optimization on 
Riemannian manifolds and in particular the method of Riemmanian Conjugate Gradient descent (RCG).

\subsection{Optimization on Riemannian Manifolds}
\label{sec:rcg}

As will be shown in \textsection~\ref{sec:proposed_method}, we need to solve a non-convex  constrained optimization problem in the form
\begin{align}
\mathrm{minimize}~\mathcal{L}(B) \notag\\
\mathrm{s.t.}~~~B \in \spd{d}\;.
\label{eqn:opt_riemannian_manifold}
\end{align}

Classical optimization methods generally turn a constrained problem into a sequence of unconstrained
problems for which unconstrained techniques can be applied.
In contrast, in this paper we make use of the optimization on Riemannian manifolds to 
minimize~\eqref{eqn:opt_riemannian_manifold}. 
This is motivated by recent advances in Riemannian optimization techniques where benefits of exploiting geometry over standard constrained optimization are shown~\cite{absil2009optimization}. As a consequence, these techniques have become increasingly popular in diverse application domains~\cite{cherian2016riemannian,harandi2015riemannian}. 

A detailed discussion of Riemannian optimization goes beyond the scope of this paper, and we refer the interested reader 
to~\cite{absil2009optimization}. 
However, the knowledge of some basic concepts will be useful in the remainder of this paper. As such, here, we briefly consider the case of Riemannian Conjugate Gradient method (RCG), our choice when the empirical study of this work is considered. First we formally define the SPD manifold.

\begin{definition}[The SPD Manifold]
The set of ($d \times d$) dimensional real, SPD matrices endowed with the Affine Invariant Riemannian Metric (AIRM)~\cite{pennec2006} forms the  SPD manifold $\spd{d}$. 
\begin{equation}
\spd{p} \triangleq \{{X} \in \mathbb{R}^{d \times d}: {v}^T {X} {v} >0,~\forall {v} \in \mathbb{R}^d-\{{0}_d\}\}\;.
\label{eqn:spd_manifold}
\end{equation}
\end{definition}

To minimize~\eqref{eqn:opt_riemannian_manifold}, RCG starts from an initial solution $\mB^{(0)}$ and improves 
its solution using the update rule
\begin{equation}
\mB^{(t+1)} = \tau_{\mB^{(t)}} 
\big( P^{(t)}
\big)\;,
\label{eqn:rcg_retraction}
\end{equation}
where  $P^{(t)}$ identifies a search direction and $\tau_{\mB} ( \cdot): T_\mB\spd{d} \to \spd{d}$
is a \emph{retraction}. The retraction serves to identify the new solution along the geodesic defined by the search direction $P^{(t)}$.
In RCG, it is guaranteed that the new solution obtained by Eq.~\eqref{eqn:rcg_retraction} is on $\spd{d}$ and has a lower objective. The search direction $P^{(t)} \in T_{\mB^{(t)}}\spd{d}$ is obtained by
\begin{equation}
P^{(t)} = -\mathrm{grad}~\mathcal{L}(\mB^{(t)}) + \eta^{(t)} \pi(P^{(t-1)},{\mB^{(t-1)} , \mB^{(t)}})\;.
\label{eqn:rcg_search_direction}
\end{equation}

Here, $\eta^{(t)}$ can be thought of as a variable learning rate, obtained via techniques such as 
Fletcher-Reeves~\cite{absil2009optimization}. Furthermore, $\mathrm{grad}~\mathcal{L}(\mB)$ is the Riemannian gradient of the objective function
at $\mB$ and $\pi(P,\mX , \mY)$ denotes the parallel transport of $P$ from $T_\mX$ to $T_\mY$. In Table~\ref{tab:riemannian_tools}, we define the mathematical entities required to perform RCG on the SPD manifold. Note that computing the standard Euclidean gradient of the function $\mathcal{L}$, denoted by $\nabla_{*}(\mathcal{L})$, is the only requirement to perform RCG on $\spd{d}$.
%

\begin{table}[t]
\small
\centering
\begin{tabular}{|c|l|}
\hline
{\bf } &{\bf \qquad\qquad\qquad $\spd{d}$}\\
\hline 
\textbf{Riemannian gradient}				&$\mathrm{grad}~\mathcal{L}(\mB) = \mB \mathrm{sym}\big(\nabla_B(\mathcal{L})\big) \mB$\\
\textbf{Retraction.}						&$\tau_\mB(\xi) = \mB^\frac{1}{2}\Exp(\mB^{-\frac{1}{2}}\xi \mB^{-\frac{1}{2}})\mB^\frac{1}{2}$\\
\textbf{Parallel Transport.}				&$\pi(P,\mX , \mY) = Z P Z^T$\\
\hline
\end{tabular}
\vspace{1ex}
\caption{Riemannian tools to perform RCG on $\spd{d}$. Here, $\mathrm{sym}(\mX) = \frac{1}{2}(\mX + \mX^T)$, 
$\Exp(\cdot)$ denotes the matrix exponential and $Z =(\mY\mX^{-{1}})^{\frac{1}{2}}$.}
\label{tab:riemannian_tools}
\end{table}

%% file: proposed_method_arxiv.tex
\section{Proposed Method}
\label{sec:proposed_method}
In this section, we first introduce the most general form of our joint $\iddl$ formulation and follow it up by providing simplifications and derivations for specific cases (such as for $\alpha=\beta=0$). 

\subsection{Information Divergence \& Dictionary Learning}
Suppose we are given a set of SPD matrices $\dataset=\set{\mX_1,\mX_2,\cdots, \mX_N},~\mX_i\in\spd{d}$ along their associated labels $\sy_i\in\labels=\set{1,2,\cdots, L}$. Our goal is three-fold: (i) learn a dictionary $\dict\in\spd{d}\times_n$, a product of $n$ SPD manifolds, (ii) learn an ABLD on each dictionary atom to best represent the given data for the task of classification, and (iii) learn a discriminative objective function on the encoded SPD matrices (in terms of $\dict$ and the respective ABLDs) for the purpose of classification.
These goals are formally captured in the $\iddl$ objective proposed below. Let the $k$-th dictionary atom in $\dict$ be $\mB_k$, then,
\begin{align}
\label{eq:iddl}
\iddl := \min_{\dict>0,\valpha>0,\vbeta>0,\mW} &   \sum_{i=1}^N  f(\vv_i, \sy_i; \mW) \\\nonumber
\text{subject to } \vv^k_i &= \abldab{\mX_i}{\mB_k}{\valpha_k}{\vbeta_k},
\end{align}
where the vector $\vv_i\in\reals{n}$ denotes the encoding of $\mX_i$ in terms of the dictionary, and $\vv_i^k$ is the $k$-th dimension of this encoding. The function $f$ parameterized by $\mW$ learns a classifier on $\vv_i$ according to the provided class labels $\sy_i$. While, there are several choices for $f$ (e.g., max-margin hinge-loss), we resort to a simple ridge regression objective in this paper. Thus, our $f$ is defined as follows: suppose $\vh_i\in\{0,1\}^n$ is a one-off encoding of class labels (i.e., $\vh_i^{\sy_i} = 1$, everywhere else zero), then
\begin{equation}
f(\vv_i, \sy_i; \mW) = \frac{1}{2}\enorm{\vh_i - \mW\vv_i}^2 + \gamma\fnorm{\mW}^2,
\label{eq:ridge}
\end{equation}
where $\mW\in\reals{L\times n}$ and $\gamma$ is a regularization parameter. Note that a separate $\alpha_k,\beta_k$ for each dictionary atom is the most general form of our formulation. In our experiments, we explore simplified cases when these parameters are shared across the atoms. 

\subsection{Efficient Optimization}
In this section, we propose efficient ways to solve the $\iddl$ objective in~\eqref{eq:iddl}. We propose to use a block-coordinate descent (BCD) scheme for optimization, in which each variable is updated alternately while fixing others. Going by the recent trends in Riemannian optimization for SPD matrices~\cite{cherian2016riemannian,harandi2015riemannian}, we use the Riemannian conjugate gradient (RCG) algorithm~\cite{absil2009optimization} for optimizing over each variable. As our objective is non-convex in its variables (except for $\mW$), convergence of BCD iterations to a global minima is not guaranteed.  In Alg.~\ref{alg:1}, we detail out the meta-steps in our optimization scheme. We initialize the dictionary atoms and the divergence parameters as described in Section~\ref{sec:param_init}. Following that, we update the atoms, the divergence parameters, and classifier parameters in an alternating manner manner -- that is, updating one variable whie fixing all others.

Recall from Section~\ref{sec:rcg} that an essential ingredient in RCG is efficient computations of the Euclidean gradients of the objective with respect to the variables. In the following, we derive expressions for these gradients. Note that we assume that the dictionary atoms (i.e., $\mB_i$) to be on an SPD manifold. 
Also w.l.o.g, we assume $\valpha$ and $\vbeta$ belong to the non-negative orthant of the Euclidean space (for reasons in Section~\ref{sec:background}).

\begin{algorithm} 
    \label{alg:1}
	\SetAlgoLined
	\KwIn{$\dataset$, $\mH$, $n$}
	$\dict \leftarrow \kmeans(\dataset, n)$, $(\valpha,\vbeta) \leftarrow \gridsearch$\;
	\Repeat{until convergence} 
	{		
		\For{$k=1$ \KwTo $n$}
		{
		    $\mB_k \leftarrow \text{update\_B}(\dataset, \mW, \valpha, \vbeta, \mB_k)$$\text{; // use \eqref{eq:21}}$
		}
		$(\valpha,\vbeta) \leftarrow \text{update\_$\alpha\beta$}(\dataset,\mW, \dict, \valpha, \vbeta)$$\text{; // use \eqref{eq:22}}$\\
		$W \leftarrow \text{update\_$\mW$}$$\text{; // using~\eqref{eq:W}}$
	}
	\KwRet{$\dict,\valpha,\vbeta$}
	\caption{Block-Coordinate Descent for IDDL.}
\end{algorithm}

\subsubsection{Gradients wrt $\mB$}
\label{sec:dict_learn}
As is clear from our formulation, only the $k$-th dimension of $\vv_i$ involves $\mB_k$. To simplify the notations, let us assume 
\begin{equation}
    \zeta = -(\vh_i - \mW\vv_i)^T\mW,
    \label{eq:zeta}
\end{equation}
and let $\zeta^k$ be its $k$-th dimension. Then we have (see the supplementary material for the details),
\begin{align}
\nabla_{\mB_k} f := \zeta^k_i \nabla_{\mB_k}\left(\abldab{\mX_i}{\mB_k}{\valpha^k}{\vbeta^k}\right).
\label{eq:20}
\end{align}
Substituting for ABLD in~\eqref{eq:20} and rearranging the terms, we have:
\begin{align}
\nabla_{\mB_k} f &= \!\frac{1}{\valpha_k\vbeta_k}\!\nabla_{\mB_k}\!\! \logdet\left[\frac{\valpha_k}{\vbeta_k}\left(\inv{\mX_i}\mB_k\right)^{\valpha_k+\vbeta_k} + \eye{d}\right] -\frac{1}{\vbeta_k}\inv{B_k}.
\label{eq:15}
\end{align}
Let $\vtheta_k = \valpha_k+\vbeta_k$ and $\vr_k = \frac{\valpha_k}{\vbeta_k}$. Further, let $\mZ_i=\inv{\mX_i}$. Then, the term inside the gradient in~\eqref{eq:15} simplifies to:
\begin{equation}
g(\mB_k; \mZ, \vr_k, \vtheta_k) = \logdet\left[\vr_k \left(\mZ\mB_k\right)^{\vtheta_k} + \eye{d}\right].
\label{eq:17}
\end{equation}

\begin{theorem}
Let $\mA,\mB \in \spd{d}$. Furthermore assume $p,q \geq 0$. We have 
\begin{align*}
&\grad{B} \logdet\left[p \left(\mA\mB\right)^{q} + \eye{d}\right] =pq \inv{\mB}\mA^{-\half}\!\!\left(\mA^{\half}\mB\mA^{\half}\right)^{q}\!\!\!\left(\eye{d} + p \left(\mA^{\half}\mB \mA^{\half}\right)^{q}\right)^{\!-1}\hspace*{-0.3cm}\mA^{\half}.
\end{align*}
\begin{proof}
For simplifying the notations, lets write $S=\mA^{\half}$. Note that the eigenvalues (and hence the $logdet$) of $\mA\mB$ is the same as that of $\mA^{\half}\mB\mA^{\half}$, however, the latter being symmetric and thus keeping $B$ symmetric when doing the gradient descent, we will use this form. Thus,
\begin{equation}
\label{eq:18}
\logdet(p\left(\mS\mB\mS\right)^{q}+\eye{d}) = \trace{\log \left(p\left(\mS\mB\mS\right)^{q}+\eye{d}\right)}
\end{equation}
Using Taylor series expansion\footnotemark, 
\begin{equation}
\label{eq:19}
\eqref{eq:18} = \trace{p (\mS\mB\mS)^{q} - \frac{p^2(\mS\mB\mS)^{2q}}{2} + \frac{p^3(\mS\mB\mS)^{3q}}{3}-\cdots}.
\end{equation}
\begin{equation}
\label{eq:20}
\!\!\grad{\mB}~\eqref{eq:19}\Rightarrow pq\mS(\mS\mB\mS)^{q-1}S - q p^2\mS(\mS\mB\mS)^{2q-1} \mS +\cdots = pq \mS\inv{(\mS\mB\mS)}\left(\mS\mB\mS\right)^{q}\left[\eye{d} - p\left(\mS\mB\mS\right)^{q}+\cdots\right]\mS.
\end{equation}
Using Maclaurin series in the middle term, we thus get $\eqref{eq:20}\Rightarrow pq\inv{\mB}\inv{\mS}\left(\mS\mB\mS\right)^{q}\inv{(\eye{d} + p \left(\mS\mB\mS\right)^{q})}\mS.$
\end{proof}
\footnotetext{Strictly speaking, the series expansions that we use in the proof assume that $p\enorm{SBS}_2\leq 1$, which can be achieved via rescaling. However, empirically this requirement has not been seen to be needed in all the datasets that we use.}
\end{theorem}

As such, the gradient $\nabla_{\mB_k} g$ is:
\begin{equation}
\grad{B_k}g\!=\!\vr_k\vtheta_k \inv{\mB_k}\mZ_i^{-\half}\!\!\left(\mZ_i^{\half}\mB_k\mZ_i^{\half}\right)^{\!\!\vtheta_k} \left(\eye{d} + \vr_k \left(\mZ_i^{\half}\mB_k \mZ_i^{\half}\right)^{\vtheta_k}\right)^{\!-1}\hspace*{-0.3cm}\mZ_i^{\half}.
\label{eq:21}
\end{equation}
Combining~\eqref{eq:21} with~\eqref{eq:15}, we have the expression for the gradient with respect to $\mB_k$.

\begin{remark}
Computing $\nabla_{\mB_k} g$ for large datasets may become overwhelming. Let $(\mU_i,\mDelta_i)$ be the Schur decomposition 
$\mZ_i^{\half}\mB_k\mZ_i^{\half}$ (which is faster than the eigenvalue decomposition~\cite{golub2012matrix}).  
With $\vdelta_i=\diag(\mDelta_i)$,  the gradient in~\eqref{eq:21} can be rewritten as:
\begin{equation}
\nabla_{\mB_k} g = \vr_k\vtheta_k\inv{\mB_k}\left(\mZ_i^{-\half}\mU_i\right) \left[\diag\left(\frac{\vdelta_i^\theta}{1+\vr_k\vdelta_i^{\theta_k}}\right)\right] \inv{\left(\mZ_i^{-\half}\mU_i\right)}.
\label{eq:gradB_simplified}
\end{equation}
Compared to~\eqref{eq:21}, this simplification reduces the number of matrix multiplications from 5 to 3 and matrix inversions from 2 to 1.
\end{remark}

\subsubsection{Gradients wrt $\valpha_k$ and $\vbeta_k$}
For gradients with respect to $\valpha_k$, we will use the form of ABLD 
given in~\eqref{eq:abld_lambda}, where $\lambda_{ijk}$ is assumed to be the $j$-th generalized eigenvalue of $\mX_i$ and dictionary atom $\mB_k$. Using the notations defined in~\eqref{eq:zeta}, the gradient has the form:
\begin{align}
    \nabla_{\valpha_k} f &= \zeta^k_i \sum_{j=1}^d \nabla_{\valpha_k} \left[\frac{1}{\valpha_k\vbeta_k} \log\frac{\valpha_k\lambda_{ijk}^{\vbeta_k} + \vbeta_k\lambda_{ijk}^{-\valpha_k}}{\valpha_k + \vbeta_k}\right] \notag \\
    &=\frac{\zeta_i^k}{\valpha_k^2\vbeta_k}\sum_{j=1}^d \Bigg\{
    \frac{\valpha_k\lambda_{ijk}^{\vbeta^k} - \valpha_k\vbeta_k\lambda_{ijk}^{-\valpha_k}\log{\lambda_{ijk}}}{\valpha_k\lambda_{ijk}^{\vbeta_k} 	    + \vbeta_k\lambda_{ijk}^{-\valpha_k}} -\frac{\valpha_k}{\valpha_k+\vbeta_k} -\log{\frac{\valpha_k\lambda_{ijk}^{\vbeta_k} + \vbeta_k\lambda_{ijk}^{-\valpha_k}}{\valpha_k+\vbeta_k}} \Bigg\}.
    \label{eq:22}
\end{align}
The gradients wrt $\vbeta_k$ from~\eqref{eq:22} can be derived using the dual symmetry property described in~\eqref{eq:dual_sym}.

\subsection{Closed Form for $\mW$}
When fixing $\dict,\valpha$ and $\vbeta$, the objective reduces to the standard ridge regression formulation in $\mW$, which can be solved in closed form as:
\begin{equation}
    \mW^* = \mH\mV^T\inv{(\mV\mV^T + \gamma\eye{d})},
    \label{eq:W}
\end{equation}
where matrices $\mV$ and $\mH$ have $\vv_i$ and $\vh_i$ along their $i$-th column, for $i=1,2,\cdots, N$.

\subsection{The Solution When $\valpha,\vbeta \rightarrow0$}
\label{sec:AIRMupdates}
As alluded to earlier, ABLD is non-smooth at the origin and we need to resort to the limit of the divergence, which happens to be the natural Riemannian metric (AIRM). That is,
\begin{equation}
    \abldab{\mX_i}{\mB_k}{0}{0} = \fnorm{\Log{\Big( \mX_i^{-\half}\mB_k\mX_i^{-\half}\Big)}}^2.
\end{equation}
Using the same ridge regression cost for $f$ defined in~\eqref{eq:ridge}, and using $\zeta_i^k$ defined in~\eqref{eq:zeta}, we have the gradient using $\mB_k$ as:
\begin{equation}
    \nabla_{\mB_k} f=2\zeta_i^k \mX_i^{-\half} \Log{\left[P_{ik}\right]}\inv{P_{ik}} \mX_i^{-\half},
    \label{eq:grad_airm}
\end{equation}
where $P_{ik} = \mX_i^{-\half}\mB_k\mX_i^{-\half}$. Note that a simplification similar to~\eqref{eq:gradB_simplified} is also possible for~\eqref{eq:grad_airm}.

\section{Computational Complexity}
We note that some of the terms in the gradients derived above could be computed offline (such as $\inv{\mX_i}$), and thus we omit those terms from our analysis. Using the simplifications depicted in~\eqref{eq:gradB_simplified} and Schur decomposition, gradient computation for each $\mB_k$ takes $\bigoh(Nd^3)$ flops. Using the gradient formulation in~\eqref{eq:22} for $\valpha$ and $\vbeta$, we need $\bigoh(Ndn + Nd^3)$ flops. Computations of the closed form for $\mW$ in~\eqref{eq:W} takes $\bigoh(n^2(L+N)+n^3+nLN)$. At test time, given that we have learned the dictionary and the parameters of the divergence, encoding a data matrix requires $\bigoh(nd^3)$ flops, which is similar in complexity to the recent sparse coding schemes such as~\cite{cherian2016riemannian}.

%% file: expts_arxiv.tex
\section{Experiments}
\label{sec:expts}
In this section, we evaluate the performance of the IDDL scheme on eight computer vision datasets, which are known to benefit from SPD-based descriptors.
To this end, we use the following datasets, namely (i) the JHMDB action recognition~\cite{jhuang2013towards}, (ii) the HMDB action recognition~\cite{kuehne2011hmdb}  (iii) the KTH-TIPS2 dataset~\cite{mallikarjuna2006kth}, (iv) Brodatz textures~\cite{ojala1996comparative}, (v) the Virus dataset~\cite{kylberg2012segmentation}, (vi) the SHREC 3D shape dataset~\cite{lai2011large}, (vii) the Myometrium cancer dataset~\cite{panos_icpr}, and (viii) the Breast cancer dataset~\cite{panos_icpr}. Below, we provide details about all the studied datasets and the way SPD descriptors are obtained on them. We use the standard evaluation schemes reported previously on these datasets. In some cases, we use our own implementations of popular methods but strictly following the recommended settings.

\subsection{Datasets}
\noindent\paragraph{HMDB and JHMDB datasets:} These are two popular action recognition benchmarks. The HMDB dataset consists of 51 action classes associated with 6766 video sequences, while JHMDB is a subset of HMDB with 955 sequences in 21 action classes. To generate SPD matrices on these datasets, we use the scheme proposed in~\cite{cherian_wacv}, where we compute RBF kernel descriptors on the output of per-frame CNN class predictions (fc8) for each stream (RBF and optical flow) separately, and fusing these two SPD matrices into a single block-diagonal matrix per sequence. For the two-stream model, we use a VGG16 model trained on optical flow and RGB frames separately as described in~\cite{simonyan2014two}. Thus, our descriptors are of size $102\times 102$ for HMDB and $42\times 42$ for JHMDB.

\noindent\paragraph{SHREC 3D Object Recognition Dataset:} It consists of 15000 RGBD covariance descriptors generated from the SHREC dataset~\cite{lai2011large} by following~\cite{fehr2013covariance}. SHREC consists of 51 3D object classes.
The descriptors are of size $18\times 18$. Similar to~\cite{cherian2016riemannian}, we randomly picked 80\% of the dataset for training and used the remaining for testing.

\noindent\paragraph{KTH-TIPS2 dataset and Brodatz Textures:} These are popular texture recognition datasets. The KTH-TIPS dataset consists of 4752 images from 11 material classes under varying conditions of illumination, pose, and scale. Covariance descriptors of size $23\times 23$ are generated from this dataset following the procedure in~\cite{harandi2014bregman}. We use the standard 4-split cross-validation for our evaluations on this dataset. As for the Brodatz dataset, we use the relative pixel coordinates, image intensity, and image gradients to form $5\times 5$ region covariance descriptors from 100 texture classes. Our dataset consists of 31000 SPD matrices, and we follow the procedure in~\cite{cherian2016riemannian} for our evaluation using an 80:20 rule as used in the RGBD dataset above.

\noindent\paragraph{Virus Dataset:}  It consists of 1500 images of 15 different virus types. Similar to the KTH-TIPS, we use the procedure in~\cite{harandi2014bregman} to generate $29\times 29$ covariance descriptors from this dataset and follow their evaluation scheme using three-splits.

\noindent\paragraph{Cancer Datasets.} Apart from these standard SPD datasets, we also report performances on two cancer recognition datasets from~\cite{panos_icpr} kindly shared with us by the authors. We use images from two types of cancers, namely (i) Breast cancer, consisting of binary classes (tissue is either cancerous or not) consisting of about 3500 samples, and (ii) Myometrium cancer, consisting of 3320 samples; we use covariance-kernel descriptors as described in~\cite{panos_icpr} which are of size $8\times 8$. We follow the 80:20 rule for evaluation on this dataset as well.

\subsection{Experimental Setup}
Since we present experiments on a variety of datasets and under various configurations, we summarize our main experiments first. There are three sets of experiments we conduct, namely (i) comparison of IDDL against other popular measures on SPD matrices, (ii) comparisons among various configurations of IDDL, and (iii) comparisons against state of the art approaches on the above datasets.  For those datasets that do not have prescribed cross-validation splits, we repeat the experiments at least 5 times and average the performance scores. 

\subsection{Parameter Initialization\label{sec:param_init}}
In all the experiments, we initialized the parameters of IDDL (e.g., the initial dictionary) in a principle-way. We initialized the dictionary atoms by applying log-Euclidean K-Means; i.e., we compute the log-Euclidean map of the SPD data, compute Euclidean K-Means on these mapped points, and remap the K-Means centroids to the SPD manifold via an exponential map. To initialize $\alpha$ and $\beta$, we recommend grid-search by fixing the dictionary atoms as above. As an alternative to the grid-search, we empirically observed that a good choice is to start with the Burg divergence (i.e., $\alpha=\beta=1$).
The regularization parameter $\lambda$ was chosen using cross-validation.

\begin{figure*}[htbp]
\begin{center}
\subfigure[JHMDB]{\label{fig:jhmdb}\includegraphics[width=5cm, trim={0cm 0cm 0cm 20cm},clip]{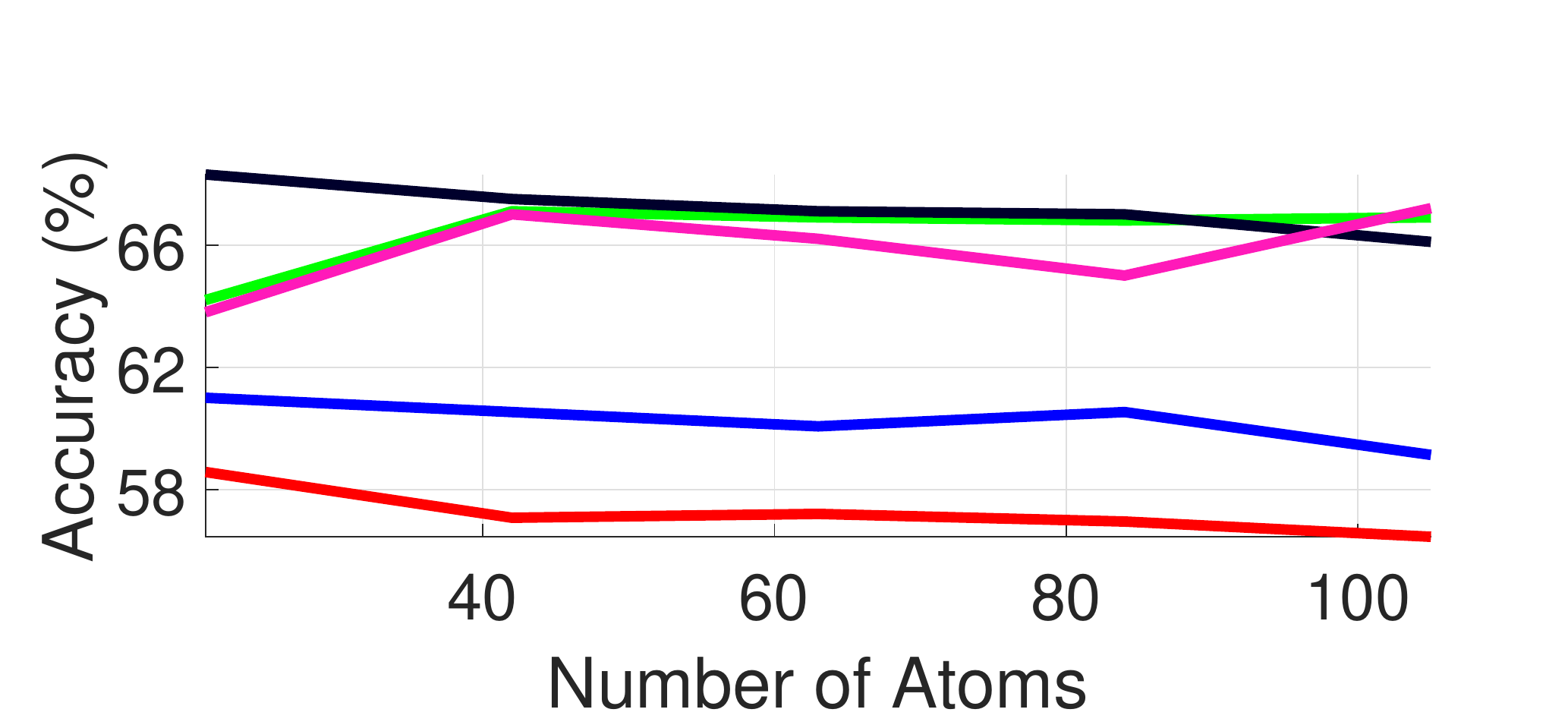}}\hspace*{0.2cm}
\subfigure[VIRUS]{\label{fig:virus}\includegraphics[width=5cm, trim={0cm 0cm 0cm 20cm},clip]{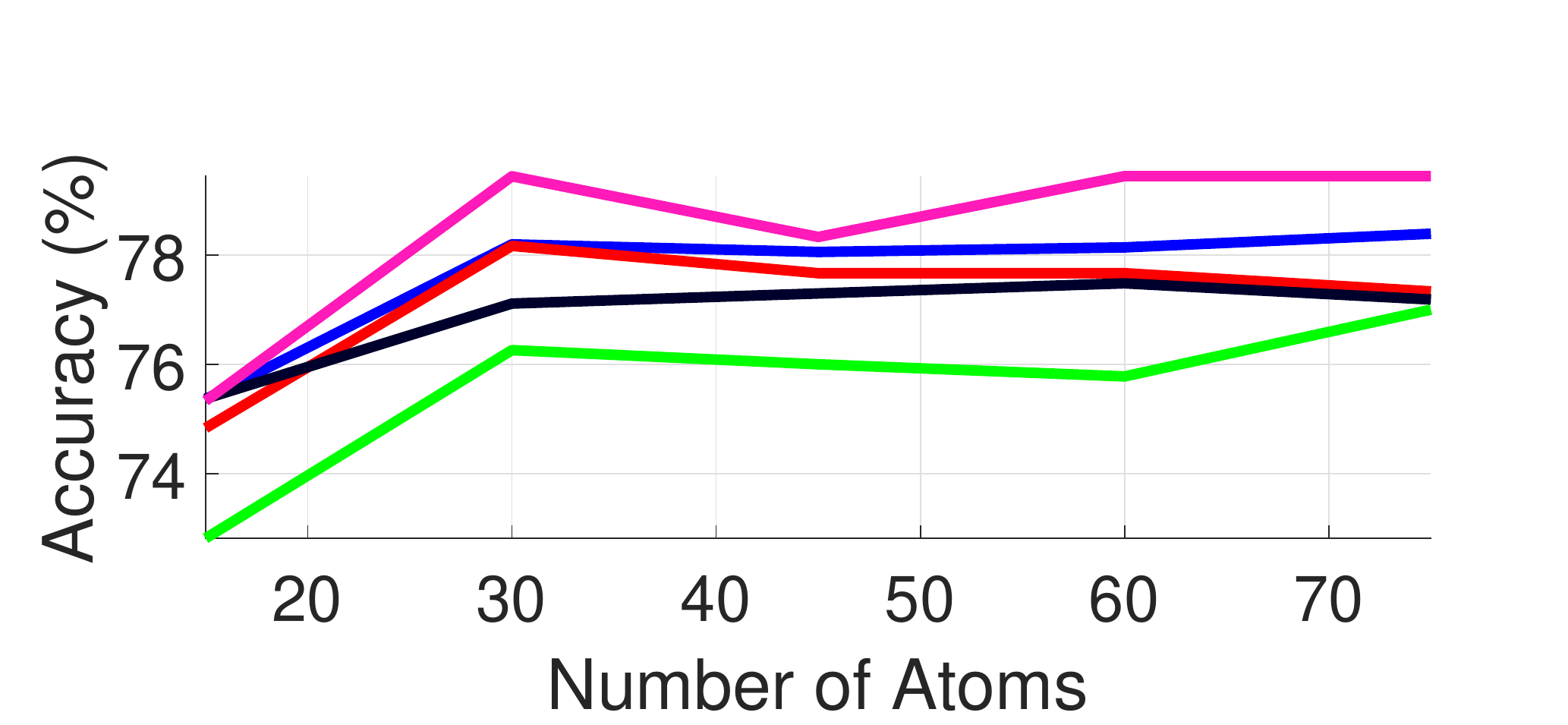}}\hspace*{0.2cm}
\subfigure[KTH-TIPS2]{\label{fig:kth}\includegraphics[width=5.7cm, trim={0cm 0cm 0cm 20cm},clip]{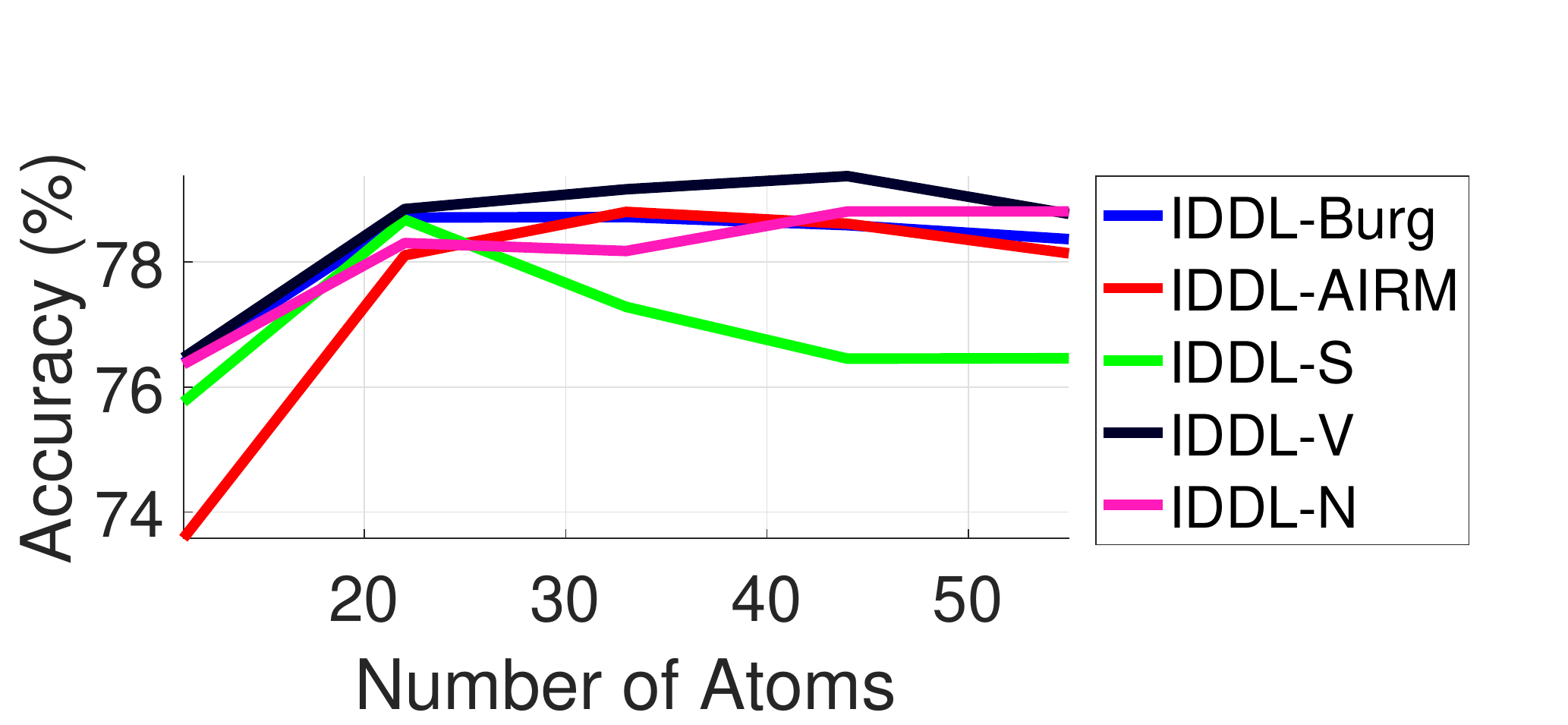}}\\
\subfigure[RGB-D Objects]{\label{fig:duc}\includegraphics[width=5cm, trim={0cm 0cm 0cm 20cm},clip]{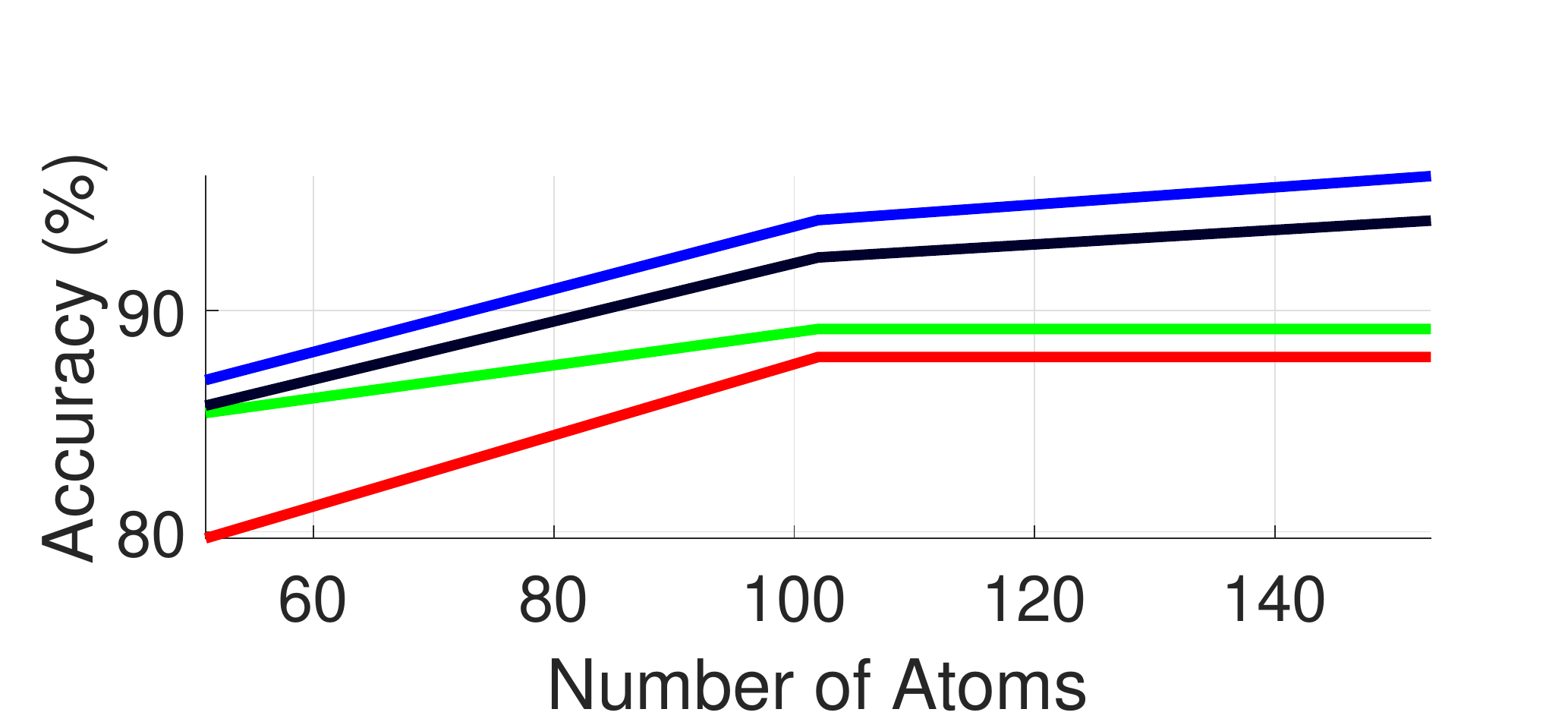}}\hspace*{0.2cm}
\subfigure[Breast Cancer]{\label{fig:bcancer}\includegraphics[width=5cm, trim={0cm 0cm 0cm 20cm},clip]{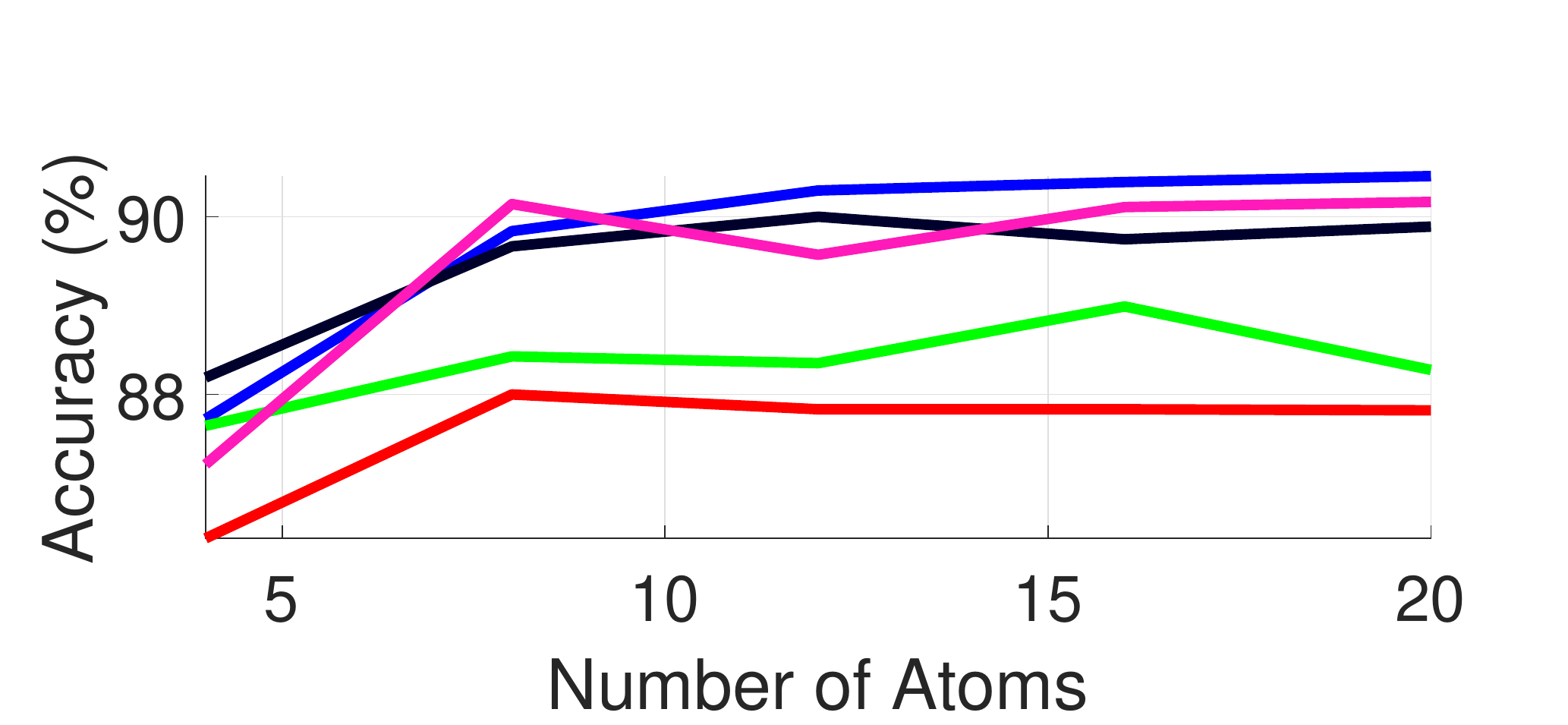}}\hspace*{0.2cm}
\subfigure[Myometrium Cancer]{\label{fig:mcancer}\includegraphics[width=5.6cm, trim={0cm 0cm 0cm 20cm},clip]{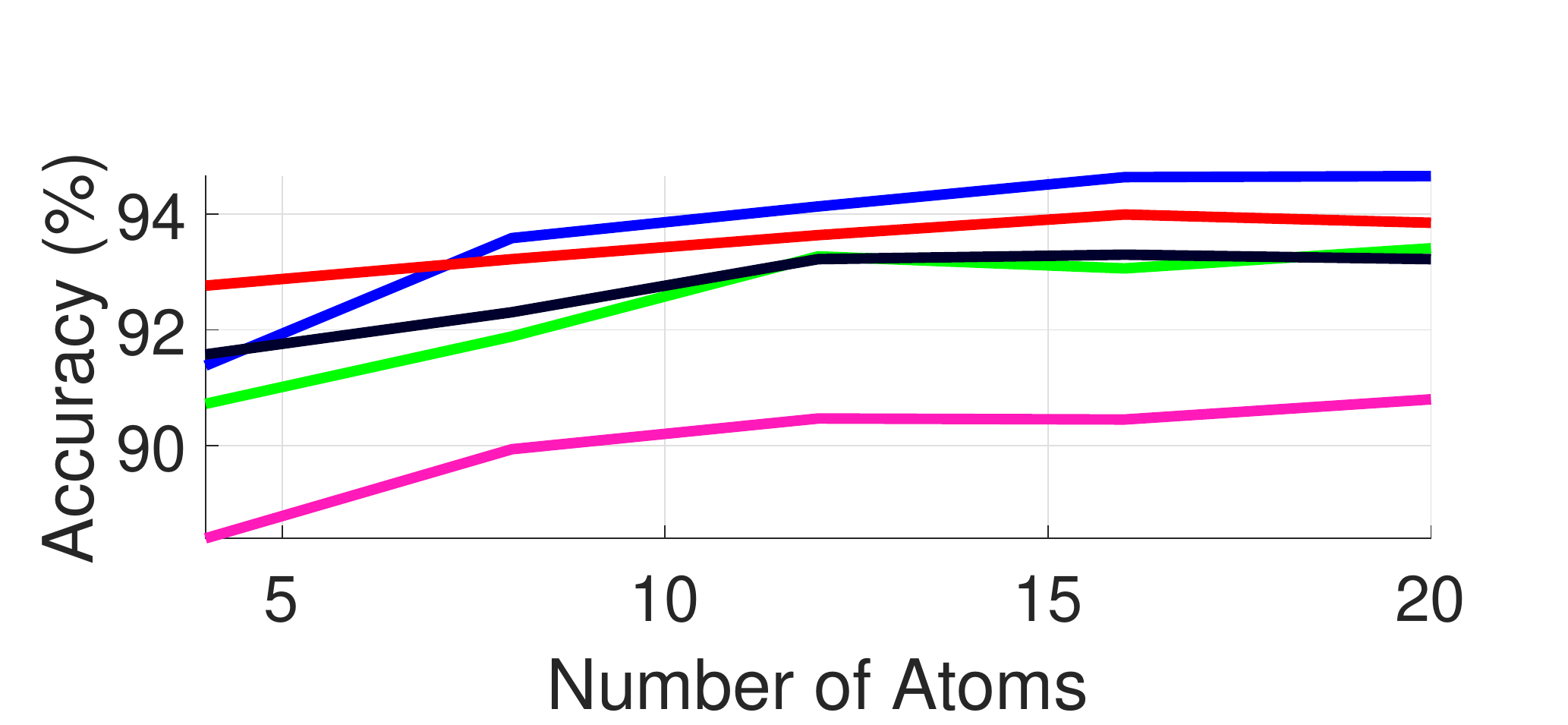}}
\caption{\label{fig:comp1st} \footnotesize{Comparisons between different variants of IDDL for increasing number of dictionary atoms.}}
\end{center}
\end{figure*}

\subsection{Comparisons to Variants of IDDL}\label{sec:variants_IDDL}
In this section, we analyze various aspects of the performance of IDDL. Generally speaking, our IDDL formulation is generic and customizable. For example, even though we formulated the problem as using a separate ABLD on each dictionary atom, it does not hurt to learn the same divergence over all atoms in some applications. To this end, we test the performance of three scenarios, namely (i) using a scalar $\alpha$ and $\beta$ that is shared across all the dictionary atoms (which we call IDDL-S), (ii) a vector $\alpha$ and $\beta$, where we assume $\alpha=\beta$, but each dictionary atom can potentially have a distinct parameter pair (we call this case IDDL-V), and (iii) the most generic case where we could have $\alpha$, $\beta$ as vectors and they may not be equal, which we refer as IDDL-N. 

In Figure~\ref{fig:comp1st}, we compare all these configurations on six of the datasets. We also include specific cases such as the Burg divergence ($\alpha=\beta=1$) and the AIRM case ($\alpha=\beta=0$) for comparisons (using the dictionary learning scheme proposed in Section~\ref{sec:dict_learn}). Our experiments show that IDDL-N and IDDL-V consistently perform well on most of the datasets. This is of course not very surprising given the generality of IDDL compared to the other measures. 

\subsection{Comparisons to Standard Measures}
In this experiment, we compare the IDDL (see Figure~\ref{fig:comp1st}) to the standard similarity measures on SPD matrices including log-Euclidean Metric~\cite{arsigny2006log}, AIRM~\cite{pennec2006}, and JBLD~\cite{cherian2013jensen}. We report 1-NN classification performance on these baselines. 
In Table~\ref{tab:1stlevel}, we report the performance of these schemes. As a rule of thumb (and also  supported empirically by cross-validation studies on our datasets), for a $C$-class problem, we chose $5C$ atoms in the dictionary. Increasing the size of the dictionary seems not helping in most cases. 
We also report a discriminative baseline by training a linear SVM on the log-Euclidean mapped SPD matrices. The results reported in Table~\ref{tab:1stlevel} clearly demonstrate the advantage of IDDL against the baselines on most of the datasets, where the benefits can go over more than 10\% in some cases (such as the JHMDB and virus).

\begin{table*}[htbp]
\centering
\begin{adjustbox}{width=1.0\linewidth}
\renewcommand{\arraystretch}{1.2} 
\begin{tabular}{c|c|c|c|c|c|c|c|c|c|c}
\small{\textbf{Dataset/Method}}  & \textbf{LEML} & \textbf{$\textrm{kSP}_{LE}$}  & \textbf{$\textrm{kSP}_{JBLD}$} & \textbf{kLLC} & \textbf{RSPDL} & \textbf{IDDL-S} & \textbf{IDDL-V} & \textbf{IDDL-N} & \textbf{IDDL-A} & \textbf{IDDL-B} \\
\hline
\textbf{JHMDB} & 58.85\% & 55.97\%  &  44.40\% & 57.46\% & 57.5\% & 67.10\% & \tb{68.3\%} & 67.20\% & 61.19\% & 61.01\%\\
\hline 
\textbf{HMDB} & 52.15\% & 44.9\% & 28.43\% & 40.20\% & 21.0\% & 52.30\% & 57.3\% & \tb{58.6\%} & 43.20\% & 46.94\%\\
\hline 
\textbf{VIRUS}& 74.60\% & 68.00\% & 57.84\% & 70.91\% & 60.8\% & 76.48\% & 77.74\% & \textbf{79.44\%} & 78.33\% & 78.37\%\\
\hline
\textbf{BRODATZ}& 47.15\% & 55.00\% & 65.19\% & 70.00\% & 74.9\% & 72.50\% & 73.2\% & 77.10\% & 62.63\% & \textbf{79.44\%}\\
\hline
\textbf{KTH TIPS} & 79.25\% & 77.18\% & 69.92\% & 73.96\% & 64.5\% & 78.68\% & 79.37\% & \textbf{79.67\%} & 78.80\% & 78.36\%\\
\hline
\textbf{3D Object} & 87.56\% & 59.26\% & 72.45\% & 87.40\% & 80.0\% & 89.17\% & 94.07\% & 92.57\% & 87.90\% & \textbf{96.08\%}\\
\hline
\textbf{Breast Cancer} & 83.18\% & 76.34\% & 71.67\% & 82.32\% & 74.2\% & 89.99\% & 90.00\% & 90.02\% & 88.00\% & \textbf{90.46\%}\\
\hline
\textbf{Myometrium Cancer} & 90.94\% & 88.69\% & 86.80\% & 88.74\% & 87.0\% & 93.41\% & 93.30\% & 90.24\% & 93.99\% & \textbf{94.66\%}\\
\end{tabular}
\end{adjustbox}
\caption{Comparisons against state of the art. IDDL-A and IDDL-B refers to IDDL-AIRM and IDDL-Burg respectively. Refer to Section~\ref{sec:variants_IDDL} for details of other abbreviations.}
\label{tab:SoA}
\end{table*}

\comment{
\begin{table}[!htbp]
\centering
\begin{tabular}{c|c|c|c|c|c|c}
\textbf{Dataset | Method}  & \textbf{LEML} & \textbf{$\textrm{kSP}_{LE}$}  & \textbf{$\textrm{kSP}_{JBLD}$}    & \textbf{kLLC} & \textbf{RSPDL} & \textbf{IDDL} & \textbf{Variant}\\
\hline
\textbf{JHMDB} & 58.85\% & 55.97\%  &  44.40\% & 57.46\% & 57.5\%& \textbf{68.3}\% &\textbf{V}\\
\hline 
\textbf{HMDB} & 52.15\% & 44.9\% & 28.43\% & 40.20\% & 21.0\%& \textbf{55.5}\% &\textbf{N} \\
\hline 
\textbf{VIRUS}& 74.60\% & 68.00\% & 57.84\% & 70.91\% & 60.8\%& \textbf{78.39}\% & \textbf{N}\\
\hline
\textbf{BRODATZ}& 47.15\% & 55.00\% & 65.19\% & 70.00\% & 74.9\% & \textbf{74.10}\% &\textbf{N}\\
\hline
\textbf{KTH TIPS} & 79.25\% & 77.18\% & 69.92\% & 73.96\% & 64.5\%& \textbf{79.37}\%& \textbf{V}\\
\hline
\textbf{3D Object} & 87.56\% & 59.26\% & 72.45\% & 87.40\% & 80.0\%& \textbf{96.08}\% & \textbf{Burg}\\
\hline
\textbf{Breast Cancer} & 83.18\% & 76.34\% & 71.67\% & 82.32\% & 74.2\%& \textbf{90.46}\% & \textbf{Burg}\\
\hline
\textbf{Myometrium Cancer} & 90.94\% & 88.69\% & 86.80\% & 88.74\% & 87.0\%& \textbf{94.66}\% &\textbf{Burg}\\
\end{tabular}
\caption{Comparisons against state of the art. Last column shows the variant of IDDL that performed the best}
\label{tab:SoA}
\end{table}
}

\subsection{Comparisons to the State of the Art}
We compare IDDL to the following popular methods that share similarities to our scheme, namely (i) Log-Euclidean Metric learning (LEML)~\cite{huang2015log}, (ii) kernelized Sparse Coding~\cite{harandi2014bregman} that uses log-Euclidean metric for sparse coding SPD matrices ($kSP_{LE}$), (iii) kernelized sparse coding using JBLD ($kSP_{JBLD}$), and kernelized locality constrained coding~\cite{harandi2015riemannian}, and Riemannian dictionary learning and sparse coding (RSPDL)~\cite{cherian2016riemannian}.  Our results are reported in Table~\ref{tab:SoA}. Again we observe that IDDL performs the best amongst all the competitive schemes, clearly demonstrating the advantage of learning the divergence and the dictionary. Note that comparisons are established by considering the same number of atoms for all schemes and fine-tuning the parameters of each algorithm (e.g., the bandwidth of the RBF kernel in $kSP_{JBLD}$) using a validation subset of the training set. 
As for LEML, we increased the number of pairwise constraints until the performance hit a plateau.


\begin{table}[!htbp]
\centering
\begin{tabular}{c|c|c|c|c|c|c}
\textbf{Dataset | Classifier}   & \textbf{LE 1-NN}  & \textbf{AIRM 1-NN}    & \textbf{JBLD 1-NN} & \textbf{SVM-LE} & \textbf{IDDL}& \textbf{Variant}\\
\hline
\textbf{JHMDB}& 52.99\% & 51.87\%& 52.24\% & 54.48\% & \textbf{68.3}\% & \textbf{V}\\
\hline
\textbf{HMDB}& 29.30\% & 43.3\%& 46.3\% & 41.7\% & \textbf{55.50}\% &\textbf{N}\\
\hline
\textbf{VIRUS}& 66.67\% & 67.89\% & 68.11\% & 68.00\% & \textbf{78.39}\% & \textbf{N}\\
\hline
\textbf{BRODATZ}& 80.10\% & 80.50\% & 80.50\% & \textbf{86.80}\% & 74.10\% & \textbf{N}\\
\hline
\textbf{KTH TIPS}& 72.05\% & 72.83\%& 72.87\% & 75.59\% & \textbf{79.37}\%& \textbf{V}\\
\hline
\textbf{3D Object}&   97.4\% & 98.2\% & 95.6\% & \textbf{98.9}\% & 96.08\%& \textbf{Burg} \\
\hline
\textbf{Breast Cancer} & 87.42\%& 80.00\% & 84.00\% & 87.71\% & \textbf{90.46}\%& \textbf{Burg}\\
\hline
\textbf{Myometrium Cancer} & 80.87\% & 84.18\% & 93.20\% & 93.22\% & \textbf{94.66}\% & \textbf{Burg}\\
\end{tabular}
\caption{Comparisons against 1-NN and SVM classification. Last column shows the variant of IDDL that worked best.}
\label{tab:1stlevel}
\end{table}

\section{Ablative Study}
\begin{figure}[!htbp]
\begin{center}
\subfigure[]{\label{fig:heatmap-b}\includegraphics[width=0.45\linewidth, trim={1cm 0cm 2cm 0.5cm},clip]{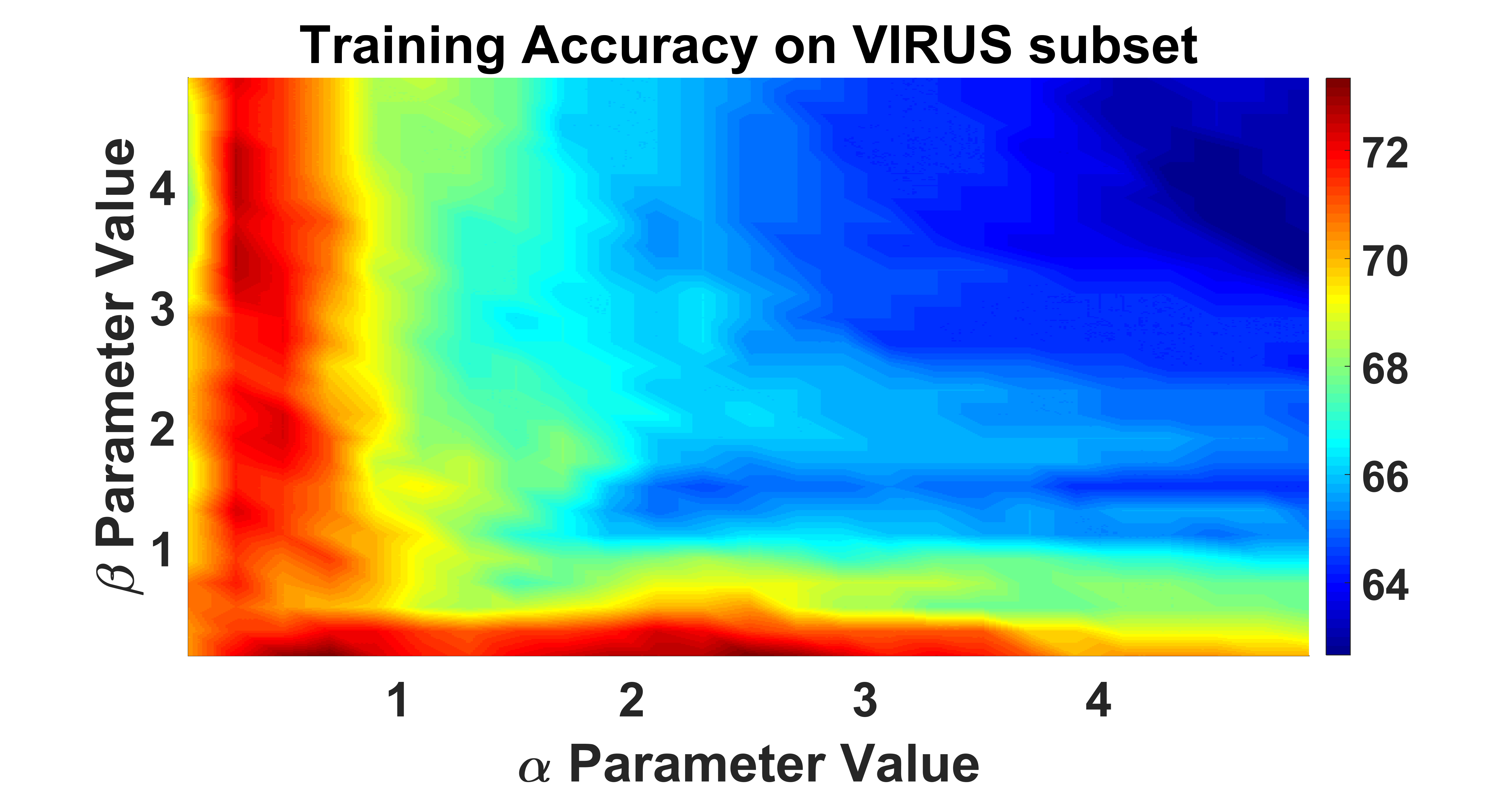}}
\caption{\label{fig:heatmap}\footnotesize{Parameter exploration for $\alpha$ and $\beta$ on (a) KTH-TIPS2 and (b) VIRUS datasets fixing the number of dictionary atoms.}}
\end{center}
\end{figure}

In this section, we study the influence of each of the components in our algorithm. In Figure~\ref{fig:heatmap}, we plot a heatmap of the classification accuracy against changing $\alpha$ and $\beta$ on the KTH-TIPS2 and Virus datasets. We fixed the size of dictionaries to 22 for the KTH TIPS and 30 for the Virus datasets. The plots  reveal that the performance varies for different parameter settings, thus (along with the results in Table~\ref{tab:1stlevel}) substantiates that learning these parameters is a way to improve performance. It should be noted that for our SVM-based experiments, we used a linear SVM on the log-Euclidean mapped SPD matrices.

\begin{figure}[!htbp]
\begin{center}
\includegraphics[width=0.45\linewidth,  trim={0cm 0cm 3cm 0.5cm},clip]{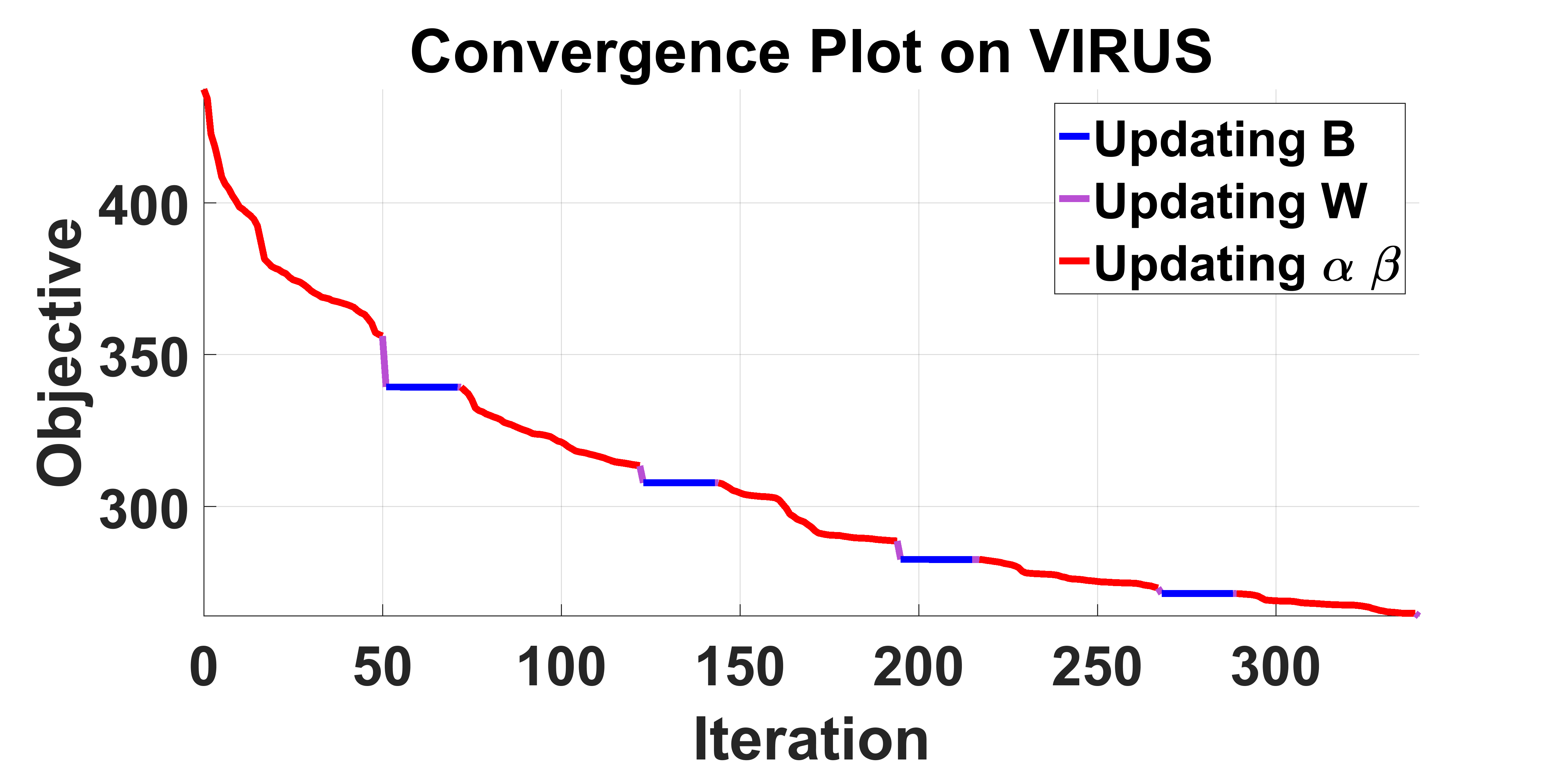}
\includegraphics[width=0.45\linewidth, trim={0cm 0cm 3cm 0.5cm},clip]{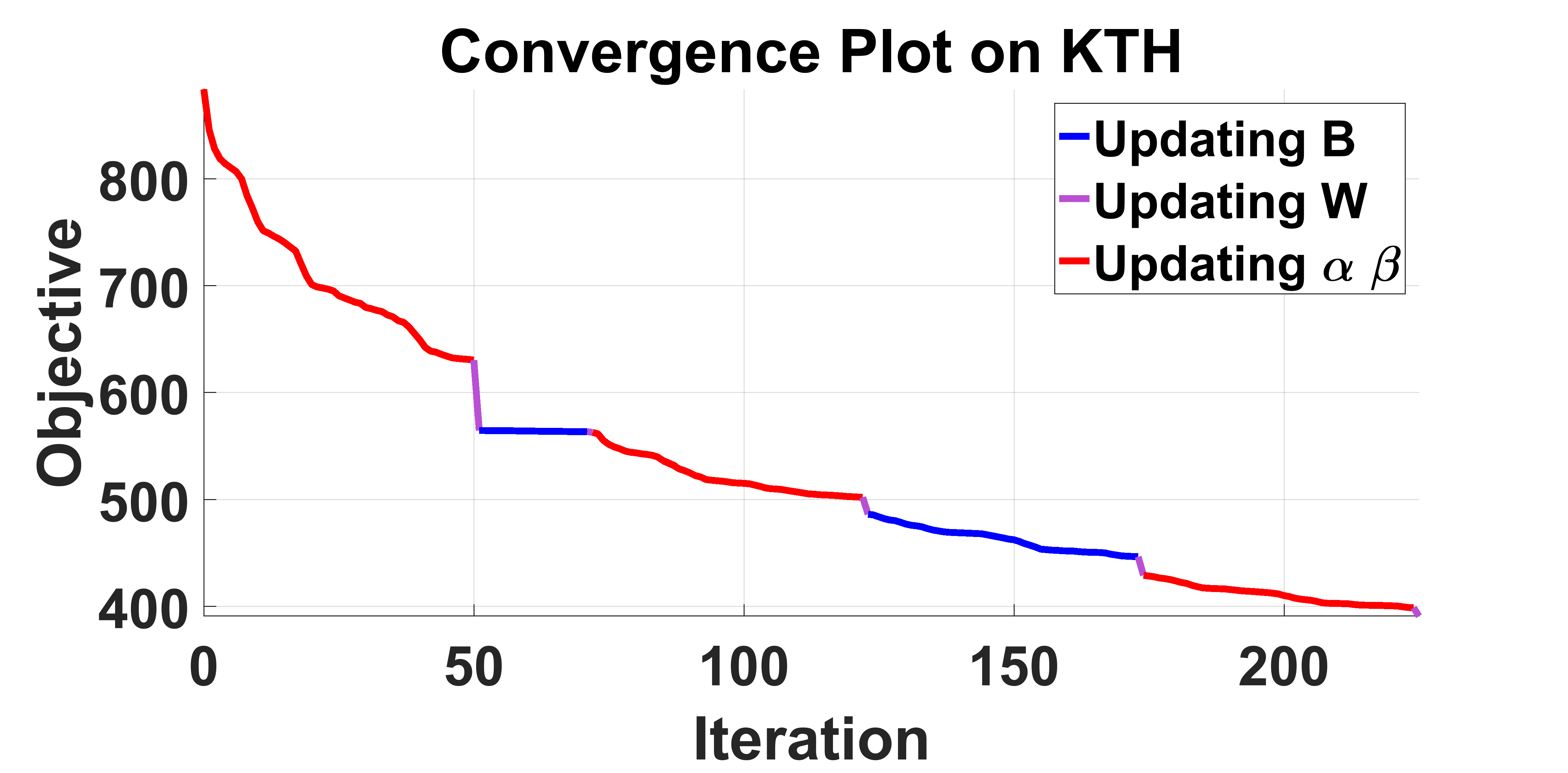}
\caption{\label{fig:convergence} \footnotesize{Convergence of the BCD optimization scheme for IDDL on the VIRUS dataset (left) and KTH-TIPS2 (right).}}
\end{center}
\end{figure}

In Figure~\ref{fig:convergence}, we plot the convergence of our objective against iterations. We also depict the BCD objective as contributed by the dictionary learning updates and the parameter learning; we use the IDDL-V for this experiment. As is clear, most part of the decrement in objective happens when the dictionary is learned, which is not surprising given that it has the most number of variables to learn. For most datasets, we observe that the RCG converges in about 200-300 iterations.



%% file: supp_mat.tex

\subsection{Running Time Experiments}
\begin{figure}[htbp]
\centering
\subfigure[]{\includegraphics[width=8cm]{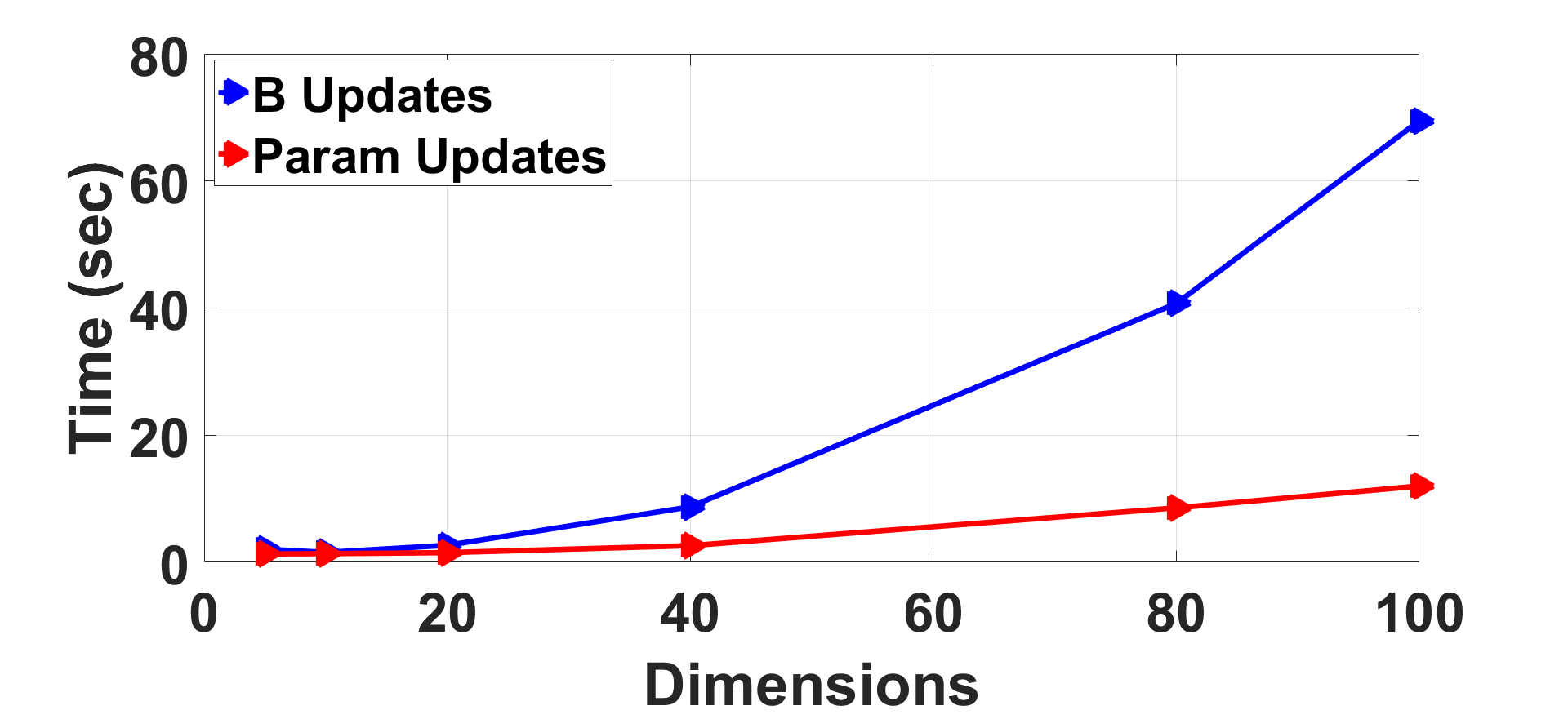}}
\subfigure[]{\includegraphics[width=8cm]{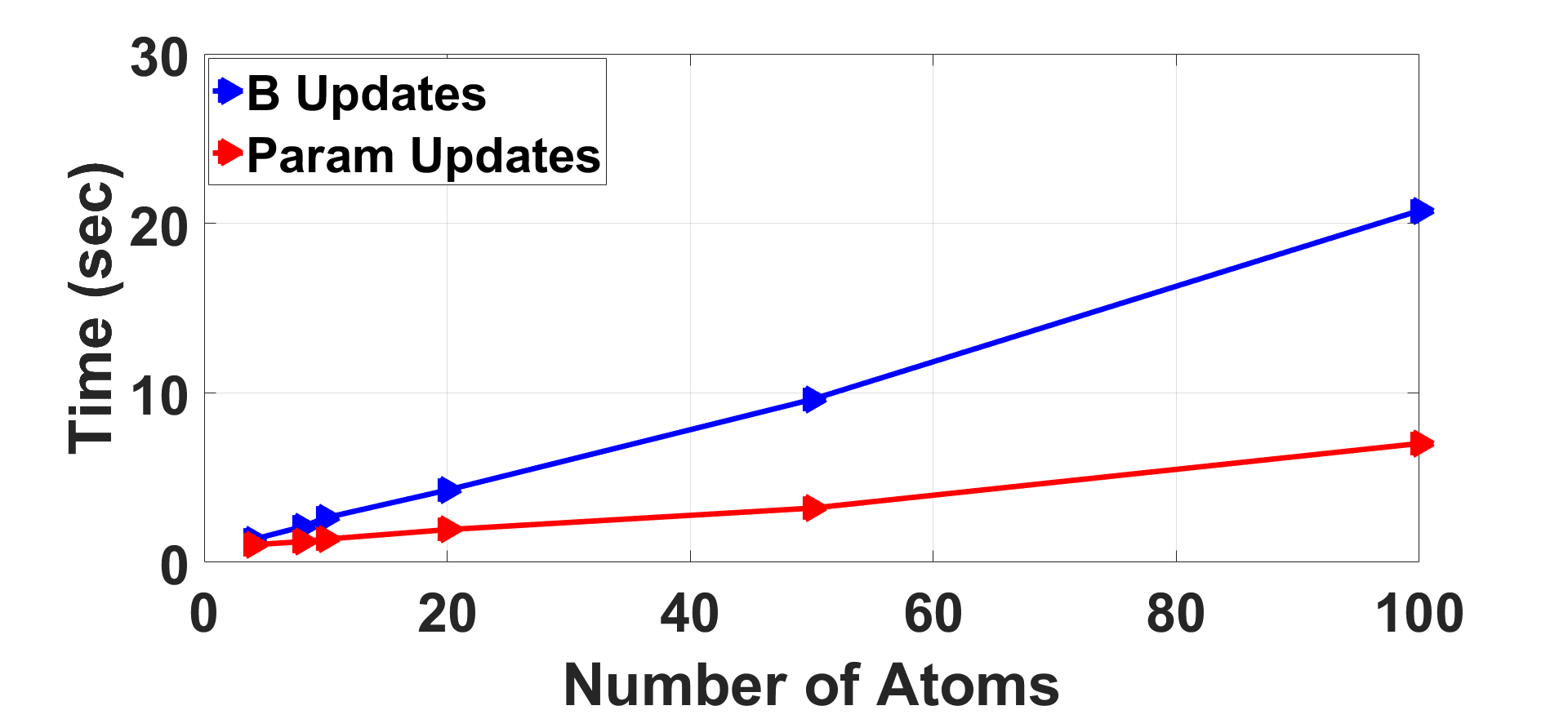}}
\caption{Time taken for one gradient computation and two objective functions evaluations against and increasing number of (a) dimensions and (b) number of atoms.}
\label{fig:timecomps}
\end{figure}

In Figure~\ref{fig:timecomps}, we plot the running time for one iteration of RCG against the number of dimensions of the matrices and the number of dictionary atoms. While our dictionary updates seem quadratic in the number of dimensions, it in fact scales linearly with the dictionary size, and usually converges in a couple of seconds.

\subsection{Evaluation of Joint Learning}
In Table~\ref{tab:paramatoms}, we evaluate the usefulness of the learning the information divergence against learning the dictionary on the Virus dataset. For this experiment, we evaluated three scenarios, (i) fixing the dictionary to the initialization (using KMeans), and learning the parameters $\alpha,\beta$ using the IDDL-S variant, (ii) fixing $\alpha,\beta$ to the initialization using GridSearch, while learning the dictionary, and (iii) learning both dictionary and the parameters jointly. As the results in Table~\ref{tab:paramatoms} shows, jointly learning the parameters demonstrates better results, thus justifying our IDDL formulation.


\begin{figure*}[htbp]
\subfigure[]{\includegraphics[width=6cm,trim={0 7cm 1cm 7cm},clip]{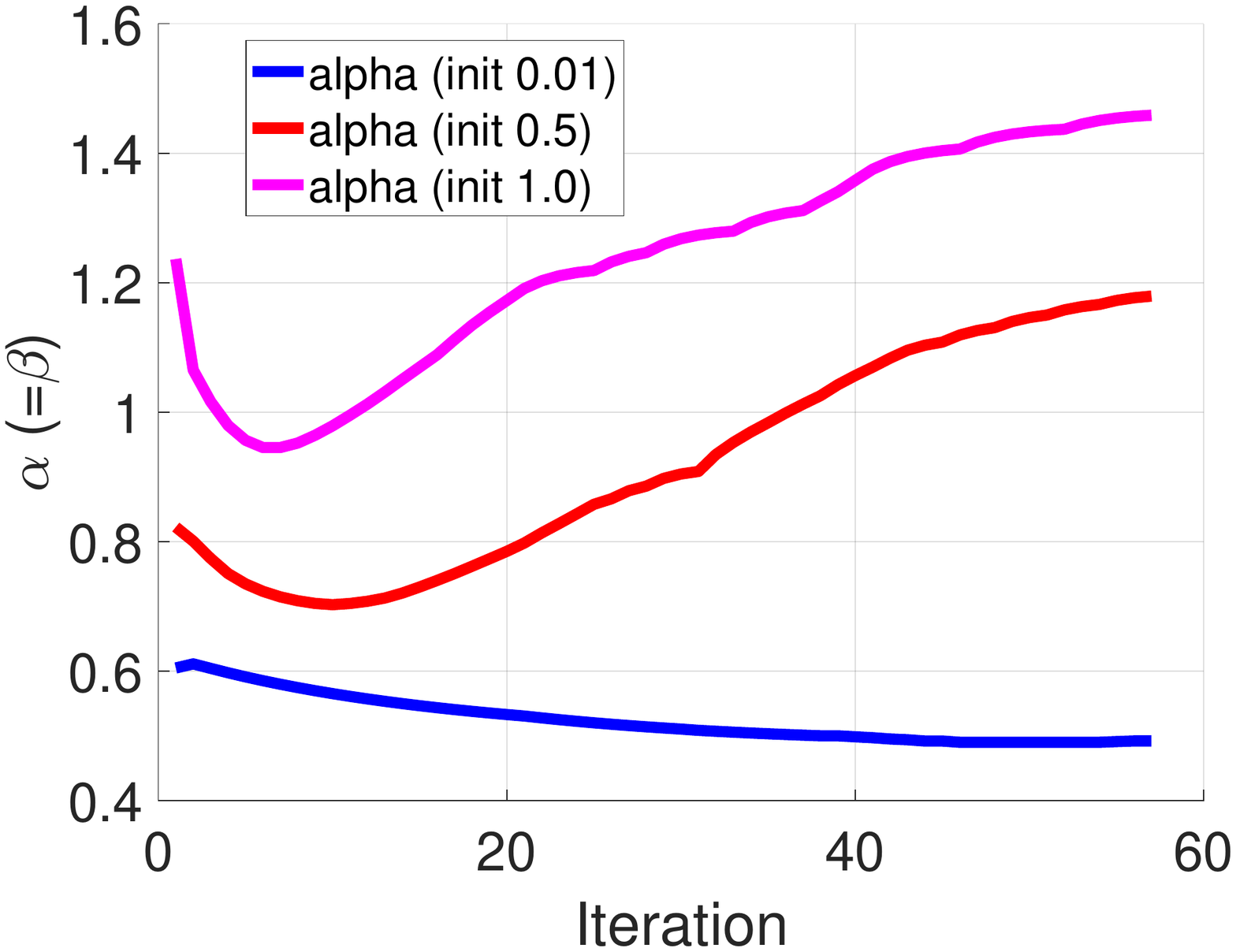}}
\subfigure[]{\includegraphics[width=6cm,trim={0 7cm 1cm 7cm},clip]{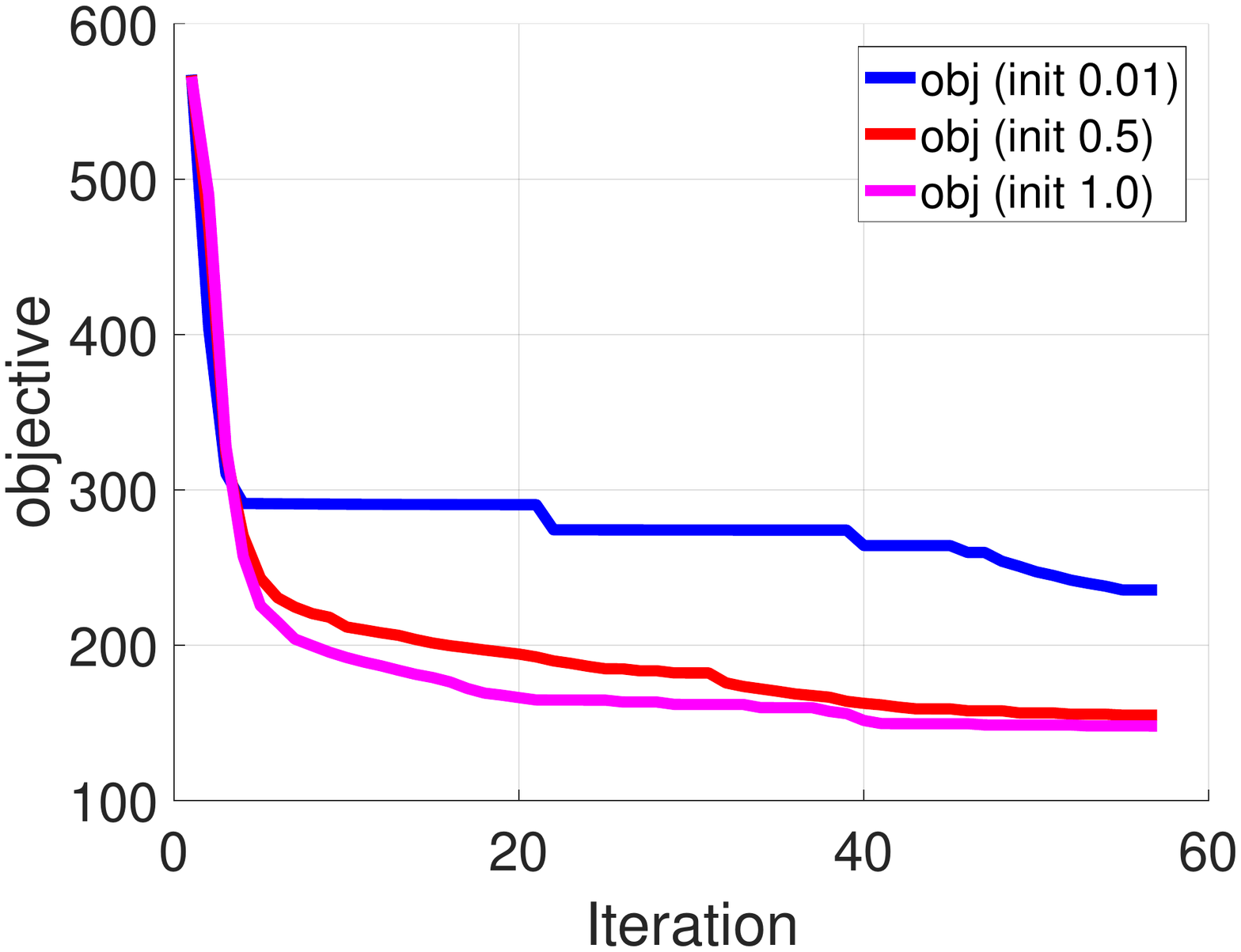}}
\subfigure[]{\includegraphics[width=6cm,trim={0 7cm 1cm 7cm},clip]{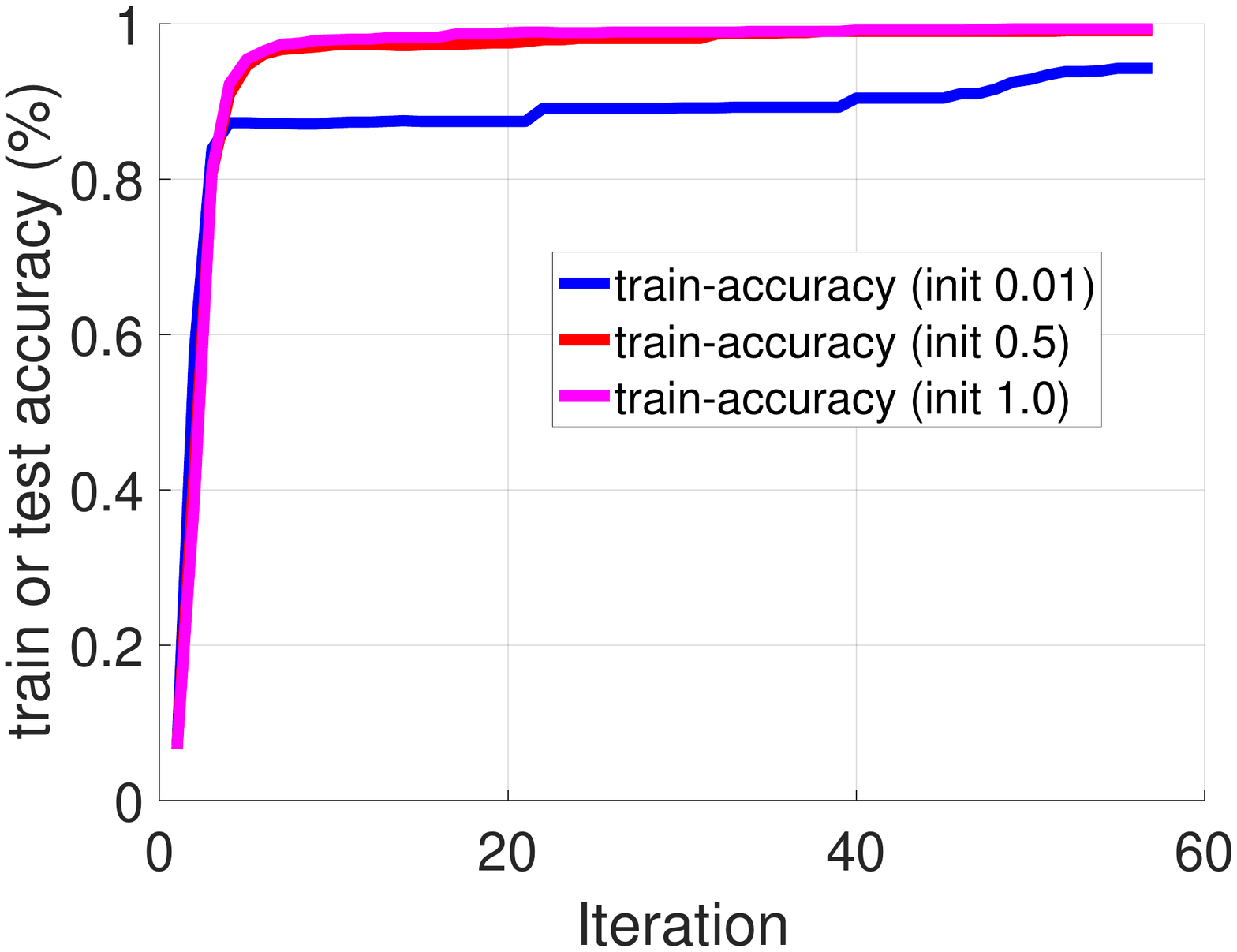}}\\
\subfigure[]{\includegraphics[width=6cm,trim={0 7cm 1cm 7cm},clip]{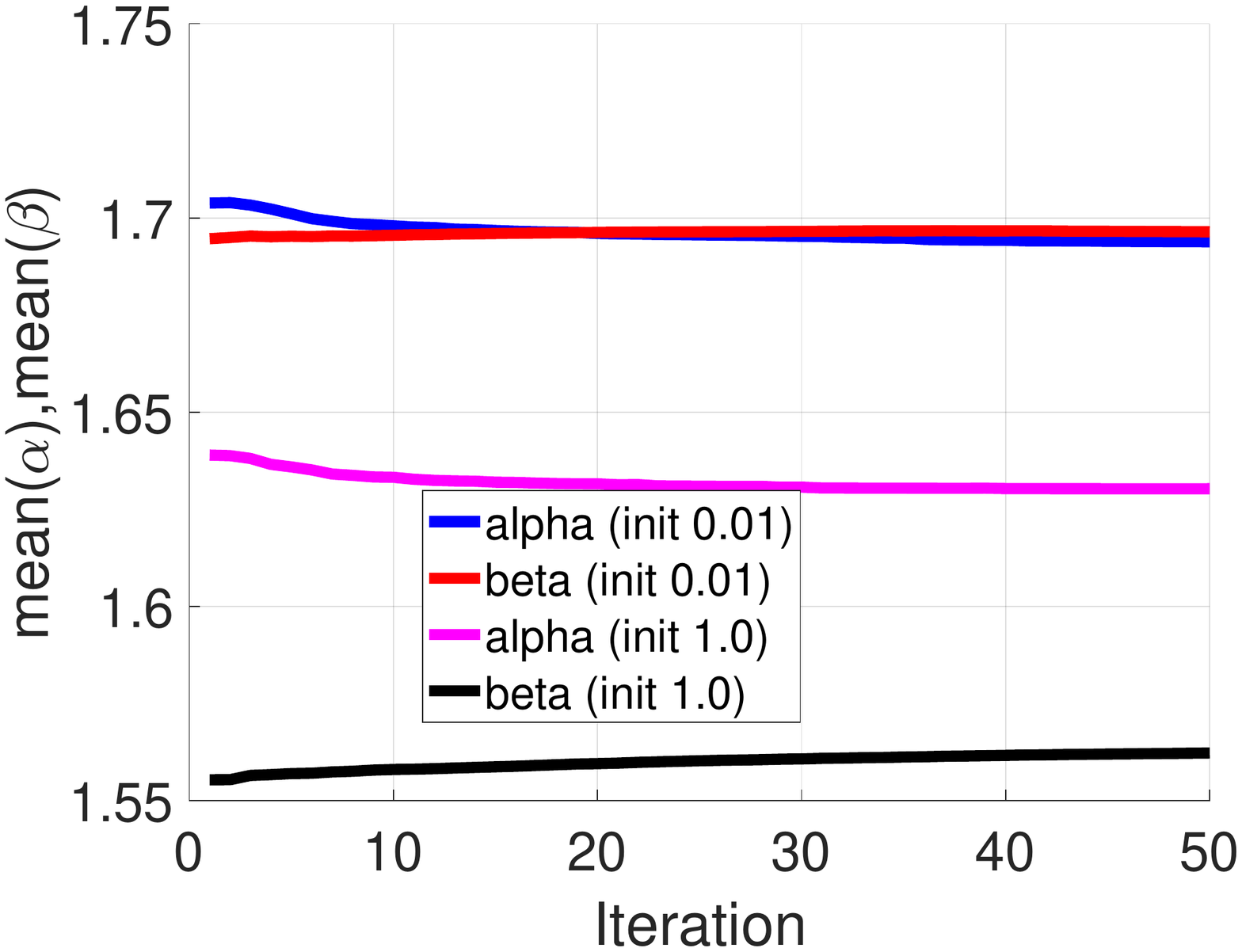}}
\subfigure[]{\includegraphics[width=6cm,trim={0 7cm 1cm 7cm},clip]{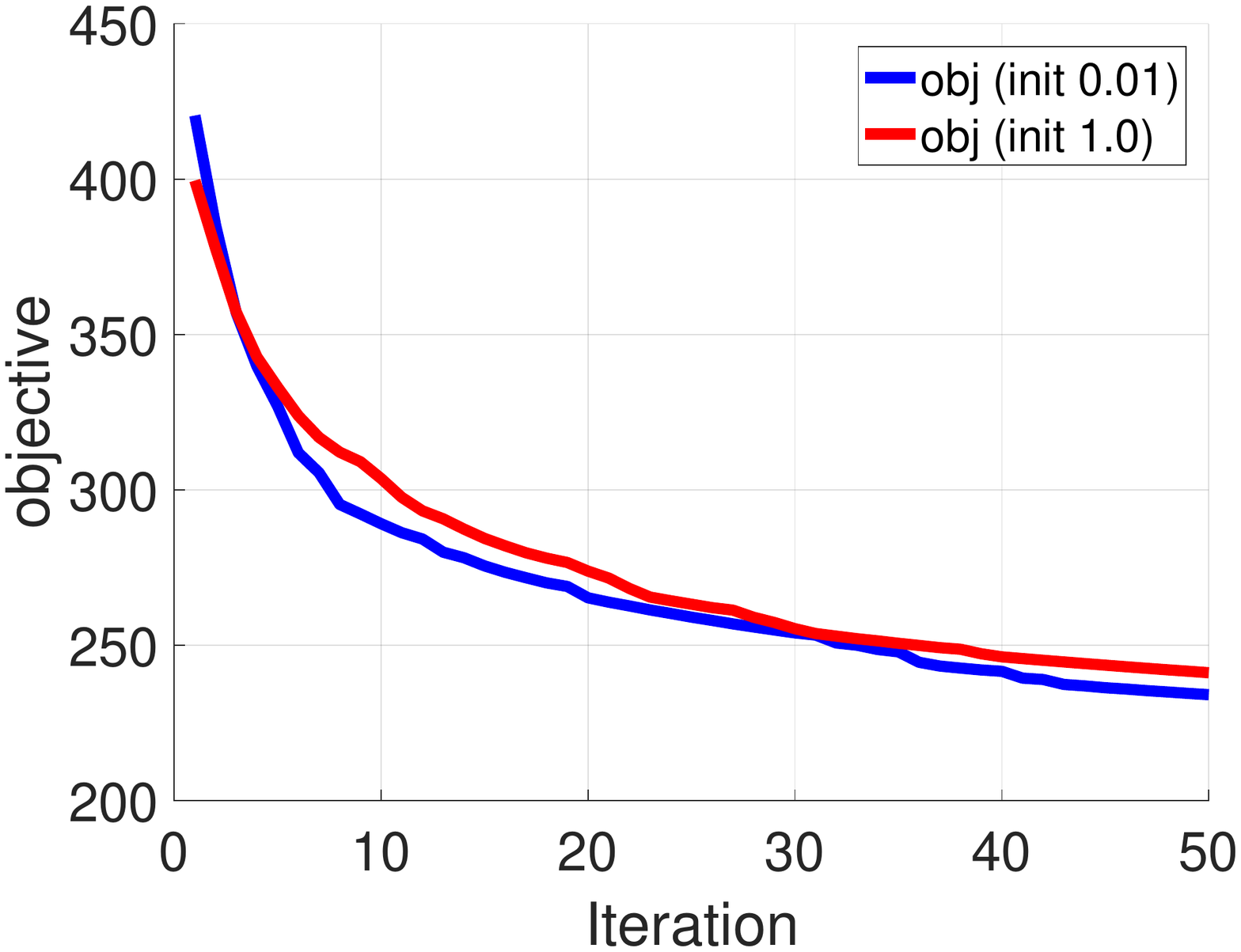}}
\subfigure[]{\includegraphics[width=6cm,trim={0 7cm 1cm 7cm},clip]{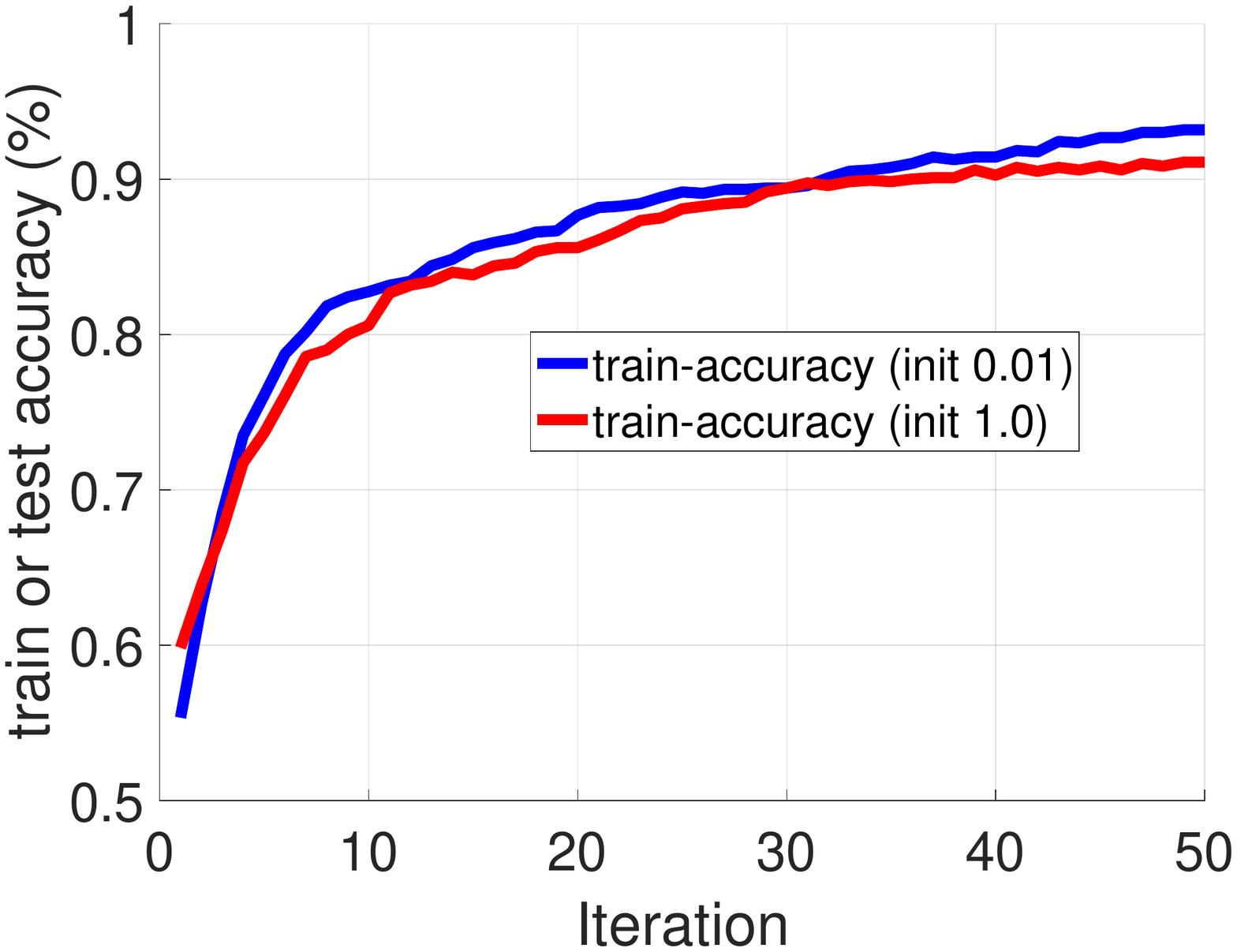}}
\caption{(a,d) Trajectory of $\alpha$ and $\beta$ over BCD iterations on the virus dataset, (b,e) shows the respective objective descent, and (c,f) shows the training set accuracy. We plot for various initializations of $\alpha$ and $\beta$.}
\label{fig:trajectory}
\end{figure*}

\begin{table}[ht]
\centering
\renewcommand{\arraystretch}{1.2} 
\begin{tabular}{c|c|c|c}
\textbf{Atoms | Method}   & \textbf{IDDL-Fix$(\alpha,\beta)$}  & \textbf{IDDL-Fix$(\mathbf{B})$} & \textbf{IDDL-N}\\
\hline
\textbf{15} & \textbf{78.33\%}& 61.67\% & 77.33\%\\
\hline
\textbf{45} & 80.33\%& 70.00\% & \textbf{83.67\%}\\
\hline
\textbf{75} & 81.67\%& 76.00\% & \textbf{82.33\%}\\
\end{tabular}
\caption{Performance evaluation of IDDL on a single split of the Virus dataset when fixing the dictionary atoms against fixing the parameters, and jointly learning the atoms and the parameters.}
\label{tab:paramatoms}
\end{table}

\subsection{Trajectories of $\alpha,\beta$}
In this experiment, we demonstrate the BCD trajectories of $\alpha$ and $\beta$ for the IDDL-S and IDDL-N variants of our algorithm on the Virus dataset. Specifically, in Figure~\ref{fig:trajectory}, we show how the value of $\alpha$ and $\beta$ varies as the BCD iteration progresses. In this experiment, we used 15 dictionary atoms. All experiments used the same initializations for the invariants. We also plot the corresponding objective and training accuracies. For IDDL-N, recall that the parameters are vectors, and thus we plot the average $\alpha$ and $\beta$ respectively. We also plot for various initializations for these parameters. 

For IDDL-S, it appears that different initializations leads to disparate points of convergence. However, for all points of convergence, the objective convergence is very similar (and so is the training accuracy), suggesting that there are multiple local minima that leads to similar empirical results. We also find that initializing with $\alpha=1.0$ demonstrates slightly better convergence than other possibilities, which we observed for other datasets too. For IDDL-N, we found that the mean of the parameters remained more or less constant, although the exact values varied by $\pm 0.2\%$ (not shown).

\section{Limitations of IDDL} 
The main limitation of our approach is the non-convexity of our objective; that precludes a formal analysis of the convergence. A further limitation is that the gradient expressions involve matrix inversions and may need careful regularizations to avoid numerical instability. We also note that the AB divergence has a discontinuity at the origin, which needs to be accounted for when learning the parameters.

Further, from our experimental analysis, it looks like there is no single variant of IDDL (amongst IDDL-S, IDDL-V, IDDL-N, IDDL-A, and IDDL-B) that consistently performs the best for all datasets. However, with the possibility of learning alpha-beta, we would think the most generalized variant IDDL-N might perhaps be the best choice for any application as it can plausibly learn the alternatives.

%% file: conclude.tex
\section{Conclusions}
\label{sec:conclude}
In this paper, we proposed a novel framework unifying the problem of dictionary learning and information divergence learning on SPD matrices; two problems that have been investigated separately so far. We leveraged on the recent advances in information geometry for this purpose, namely using the $\alpha\beta$-logdet divergence. We formulated an objective for jointly learning the divergence and the dictionary and showed that it can be solved efficiently using optimization methods on Riemannian manifolds. Experiments on eight computer vision datasets demonstrate superior performance of our approach against alternatives. 

%% file: ablearn_arxiv.bbl
\begin{thebibliography}{10}\itemsep=-1pt

\bibitem{absil2009optimization}
P.-A. Absil, R.~Mahony, and R.~Sepulchre.
\newblock {\em Optimization algorithms on matrix manifolds}.
\newblock Princeton University Press, 2009.

\bibitem{amari2007methods}
S.-i. Amari and H.~Nagaoka.
\newblock {\em Methods of information geometry}, volume 191.
\newblock American Mathematical Soc., 2007.

\bibitem{arsigny2006log}
V.~Arsigny, P.~Fillard, X.~Pennec, and N.~Ayache.
\newblock Log-euclidean metrics for fast and simple calculus on diffusion
  tensors.
\newblock {\em Magnetic resonance in medicine}, 56(2):411--421, 2006.

\bibitem{basu1998robust}
A.~Basu, I.~R. Harris, N.~L. Hjort, and M.~Jones.
\newblock Robust and efficient estimation by minimising a density power
  divergence.
\newblock {\em Biometrika}, 85(3):549--559, 1998.

\bibitem{brox2006nonlinear}
T.~Brox, J.~Weickert, B.~Burgeth, and P.~Mr{\'a}zek.
\newblock Nonlinear structure tensors.
\newblock {\em Image and Vision Computing}, 24(1):41--55, 2006.

\bibitem{carreira2012semantic}
J.~Carreira, R.~Caseiro, J.~Batista, and C.~Sminchisescu.
\newblock Semantic segmentation with second-order pooling.
\newblock In {\em ECCV}, 2012.

\bibitem{cherian_wacv}
A.~Cherian, P.~Koniusz, and S.~Gould.
\newblock Higher-order pooling of {CNN} features via kernel linearization for
  action recognition.
\newblock In {\em WACV}, 2017.

\bibitem{cherian2016riemannian}
A.~Cherian and S.~Sra.
\newblock Riemannian dictionary learning and sparse coding for positive
  definite matrices.
\newblock {\em IEEE Trans. on Neural Networks and Learning Systems}, 2016.

\bibitem{cherian2013jensen}
A.~Cherian, S.~Sra, A.~Banerjee, and N.~Papanikolopoulos.
\newblock Jensen-bregman logdet divergence with application to efficient
  similarity search for covariance matrices.
\newblock {\em PAMI}, 35(9):2161--2174, 2013.

\bibitem{cichocki2010families}
A.~Cichocki and S.-i. Amari.
\newblock Families of alpha-beta-and gamma-divergences: Flexible and robust
  measures of similarities.
\newblock {\em Entropy}, 12(6):1532--1568, 2010.

\bibitem{cichocki2015log}
A.~Cichocki, S.~Cruces, and S.-i. Amari.
\newblock Log-determinant divergences revisited: Alpha-beta and gamma log-det
  divergences.
\newblock {\em Entropy}, 17(5):2988--3034, 2015.

\bibitem{cichocki2009nonnegative}
A.~Cichocki, R.~Zdunek, A.~H. Phan, and S.-i. Amari.
\newblock {\em Non-negative matrix and tensor factorizations: applications to
  exploratory multi-way data analysis and blind source separation}.
\newblock John Wiley \& Sons, 2009.

\bibitem{dhillon2005generalized}
I.~S. Dhillon and S.~Sra.
\newblock Generalized nonnegative matrix approximations with bregman
  divergences.
\newblock In {\em NIPS}, 2005.

\bibitem{dikmen2015learning}
O.~Dikmen, Z.~Yang, and E.~Oja.
\newblock Learning the information divergence.
\newblock {\em PAMI}, 37(7):1442--1454, 2015.

\bibitem{fehr2013covariance}
D.~Fehr.
\newblock {\em Covariance based point cloud descriptors for object detection
  and classification}.
\newblock PhD thesis, University Of Minnesota, 2013.

\bibitem{golub2012matrix}
G.~H. Golub and C.~F. Van~Loan.
\newblock {\em Matrix computations}, volume~3.
\newblock JHU Press, 2012.

\bibitem{harandi2015riemannian}
M.~Harandi and M.~Salzmann.
\newblock Riemannian coding and dictionary learning: Kernels to the rescue.
\newblock In {\em CVPR}, 2015.

\bibitem{harandi2014bregman}
M.~Harandi, M.~Salzmann, and F.~Porikli.
\newblock Bregman divergences for infinite dimensional covariance matrices.
\newblock In {\em CVPR}, 2014.

\bibitem{harandi2014manifold}
M.~T. Harandi, M.~Salzmann, and R.~Hartley.
\newblock From manifold to manifold: Geometry-aware dimensionality reduction
  for spd matrices.
\newblock In {\em ECCV}, 2014.

\bibitem{hinton2002stochastic}
G.~Hinton and S.~Roweis.
\newblock Stochastic neighbor embedding.
\newblock In {\em NIPS}, 2002.

\bibitem{huang2016riemannian}
Z.~Huang and L.~Van~Gool.
\newblock A {R}iemannian network for {SPD} matrix learning.
\newblock {\em CoRR arXiv:1608.04233}, 2016.

\bibitem{huang2015log}
Z.~Huang, R.~Wang, S.~Shan, X.~Li, and X.~Chen.
\newblock {Log-Euclidean} metric learning on symmetric positive definite
  manifold with application to image set classification.
\newblock In {\em ICML}, 2015.

\bibitem{ionescu2015matrix}
C.~Ionescu, O.~Vantzos, and C.~Sminchisescu.
\newblock Matrix backpropagation for deep networks with structured layers.
\newblock In {\em ICCV}, 2015.

\bibitem{jhuang2013towards}
H.~Jhuang, J.~Gall, S.~Zuffi, C.~Schmid, and M.~J. Black.
\newblock Towards understanding action recognition.
\newblock In {\em ICCV}, 2013.

\bibitem{kompass2007generalized}
R.~Kompass.
\newblock A generalized divergence measure for nonnegative matrix
  factorization.
\newblock {\em Neural computation}, 19(3):780--791, 2007.

\bibitem{kuehne2011hmdb}
H.~Kuehne, H.~Jhuang, E.~Garrote, T.~Poggio, and T.~Serre.
\newblock Hmdb: a large video database for human motion recognition.
\newblock In {\em ICCV}, 2011.

\bibitem{kulis2006learning}
B.~Kulis, M.~Sustik, and I.~Dhillon.
\newblock Learning low-rank kernel matrices.
\newblock In {\em ICML}, 2006.

\bibitem{kylberg2012segmentation}
G.~Kylberg, M.~Uppstr{\"o}m, K.~Hedlund, G.~Borgefors, and I.~Sintorn.
\newblock Segmentation of virus particle candidates in transmission electron
  microscopy images.
\newblock {\em Journal of microscopy}, 245(2):140--147, 2012.

\bibitem{lafferty1999additive}
J.~Lafferty.
\newblock Additive models, boosting, and inference for generalized divergences.
\newblock In {\em Proc. conf. on Computational learning theory}, 1999.

\bibitem{lai2011large}
K.~Lai, L.~Bo, X.~Ren, and D.~Fox.
\newblock A large-scale hierarchical multi-view {RGB-D} object dataset.
\newblock In {\em ICRA}, 2011.

\bibitem{li2013log}
P.~Li, Q.~Wang, W.~Zuo, and L.~Zhang.
\newblock {Log-Euclidean} kernels for sparse representation and dictionary
  learning.
\newblock In {\em ICCV}, 2013.

\bibitem{mallikarjuna2006kth}
P.~Mallikarjuna, A.~T. Targhi, M.~Fritz, E.~Hayman, B.~Caputo, and J.-O.
  Eklundh.
\newblock The {KTH-TIPS2} database, 2006.

\bibitem{mihoko2002robust}
M.~Mihoko and S.~Eguchi.
\newblock Robust blind source separation by beta divergence.
\newblock {\em Neural computation}, 14(8):1859--1886, 2002.

\bibitem{moakher2006symmetric}
M.~Moakher and P.~G. Batchelor.
\newblock Symmetric positive-definite matrices: From geometry to applications
  and visualization.
\newblock In {\em Visualization and Processing of Tensor Fields}, pages
  285--298. Springer, 2006.

\bibitem{ojala1996comparative}
T.~Ojala, M.~Pietik{\"a}inen, and D.~Harwood.
\newblock A comparative study of texture measures with classification based on
  featured distributions.
\newblock {\em Pattern recognition}, 29(1):51--59, 1996.

\bibitem{pennec2006}
X.~Pennec, P.~Fillard, and N.~Ayache.
\newblock A {Riemannian} framework for tensor computing.
\newblock {\em IJCV}, 66(1):41--66, 2006.

\bibitem{simonyan2014two}
K.~Simonyan and A.~Zisserman.
\newblock Two-stream convolutional networks for action recognition in videos.
\newblock In {\em NIPS}, 2014.

\bibitem{csimcsekli2015learning}
U.~{\c{S}}im{\c{s}}ekli, A.~T. Cemgil, and B.~Ermi{\c{s}}.
\newblock Learning mixed divergences in coupled matrix and tensor factorization
  models.
\newblock In {\em ICASSP}, 2015.

\bibitem{sivalingam2010tensor}
R.~Sivalingam, D.~Boley, V.~Morellas, and N.~Papanikolopoulos.
\newblock Tensor sparse coding for region covariances.
\newblock In {\em ECCV}, 2010.

\bibitem{sivalingam2009metric}
R.~Sivalingam, V.~Morellas, D.~Boley, and N.~Papanikolopoulos.
\newblock Metric learning for semi-supervised clustering of region covariance
  descriptors.
\newblock In {\em ICDSC}, 2009.

\bibitem{panos_icpr}
P.~Stanitsas, A.~Cherian, X.~Li, A.~Truskinovsky, V.~Morellas, and
  N.~Papanikolopoulos.
\newblock Evaluation of feature descriptors for cancerous tissue recognition.
\newblock In {\em ICPR}, 2016.

\bibitem{thiyam2017optimization}
D.~B. Thiyam, S.~Cruces, J.~Olias, and A.~Cichocki.
\newblock Optimization of alpha-beta log-det divergences and their application
  in the spatial filtering of two class motor imagery movements.
\newblock {\em Entropy}, 19(3):89, 2017.

\bibitem{tuzel2006region}
O.~Tuzel, F.~Porikli, and P.~Meer.
\newblock Region covariance: A fast descriptor for detection and
  classification.
\newblock In {\em ECCV}, 2006.

\bibitem{wang2015beyond}
L.~Wang, J.~Zhang, L.~Zhou, C.~Tang, and W.~Li.
\newblock Beyond covariance: Feature representation with nonlinear kernel
  matrices.
\newblock In {\em ICCV}, 2015.

\bibitem{wang2012covariance}
R.~Wang, H.~Guo, L.~S. Davis, and Q.~Dai.
\newblock Covariance discriminative learning: A natural and efficient approach
  to image set classification.
\newblock In {\em CVPR}, 2012.

\bibitem{xie2013nonlinear}
Y.~Xie, J.~Ho, and B.~Vemuri.
\newblock On a nonlinear generalization of sparse coding and dictionary
  learning.
\newblock In {\em ICML}, 2013.

\end{thebibliography}
